%% file: FullManuscript.tex
\documentclass{article}

\input{PackagesMacros}

\icmltitlerunning{\myTitle}

\begin{document}

\twocolumn[
\icmltitle{\myTitle}



\icmlsetsymbol{equal}{*}

\begin{icmlauthorlist}
\icmlauthor{Vladimir Kulikov}{yyy}
\icmlauthor{Shahar Yadin}{equal,yyy}
\icmlauthor{Matan Kleiner}{equal,yyy}
\icmlauthor{Tomer Michaeli}{yyy}
\end{icmlauthorlist}

\icmlaffiliation{yyy}{Faculty of Electrical and Computer Engineering, Technion – Israel Institute of Technology, Haifa, Israel.}

\icmlcorrespondingauthor{Vladimir Kulikov}{vladimir.k@campus.technion.ac.il}

\icmlkeywords{Machine Learning, ICML, Generative Model, Single Image, Diffusion Model, Text Guided Image Generation}

\vskip 0.3in
]



\printAffiliationsAndNotice{\icmlEqualContribution} 

\input{Abstract}

\input{Intro}

\input{RelatedWork}

\input{Method}

\input{Experiments}

\input{Conclusion}

\input{Acknowledgments}

\bibliography{SinDDMRef}
\bibliographystyle{icml2023}



\clearpage
\onecolumn
\appendix
\renewcommand{\thefigure}{S\arabic{figure}} 
\renewcommand{\theHfigure}{S\arabic{figure}}
\setcounter{figure}{0}  
\renewcommand\theequation{S\arabic{equation}} 
\renewcommand{\theHequation}{S\arabic{figure}}
\setcounter{equation}{0}  
\renewcommand{\thesection}{\Alph{section}}

\input{Appendix}

\end{document}

%% file: PackagesMacros.tex
\PassOptionsToPackage{dvipsnames}{xcolor}

\usepackage{microtype}
\usepackage{graphicx}
\usepackage{subfigure}
\usepackage{booktabs} 

\usepackage[accepted]{icml2023}

\usepackage{amsmath}
\usepackage{amssymb}
\usepackage{multirow}
\usepackage{xr}
\usepackage{xr-hyper}
\usepackage{hyperref}

\graphicspath{ {./images/} }

\usepackage[capitalize,noabbrev]{cleveref}
\crefname{section}{Sec.}{Secs.}
\Crefname{section}{Section}{Sections}
\Crefname{table}{Table}{Tables}
\crefname{table}{Tab.}{Tabs.}

\hypersetup{
	hidelinks,
	colorlinks=true,
	linkcolor=Blue,
	filecolor=Blue,
	citecolor=Blue,
	urlcolor=Blue,
	breaklinks=false
}

\newcommand{\myTitle}{SinDDM: A Single Image Denoising Diffusion Model}

\makeatletter
\newcommand*{\addFileDependency}[1]{
	\typeout{(#1)}
	\@addtofilelist{#1}
	\IfFileExists{#1}{\typeout{File #1 O.K.}}{\typeout{No file #1.}}
}
\makeatother

\newcommand{\eg}{\emph{e.g. }}



%% file: Abstract.tex
\begin{figure*}[h]
    \centering
    \includegraphics[width=\linewidth]{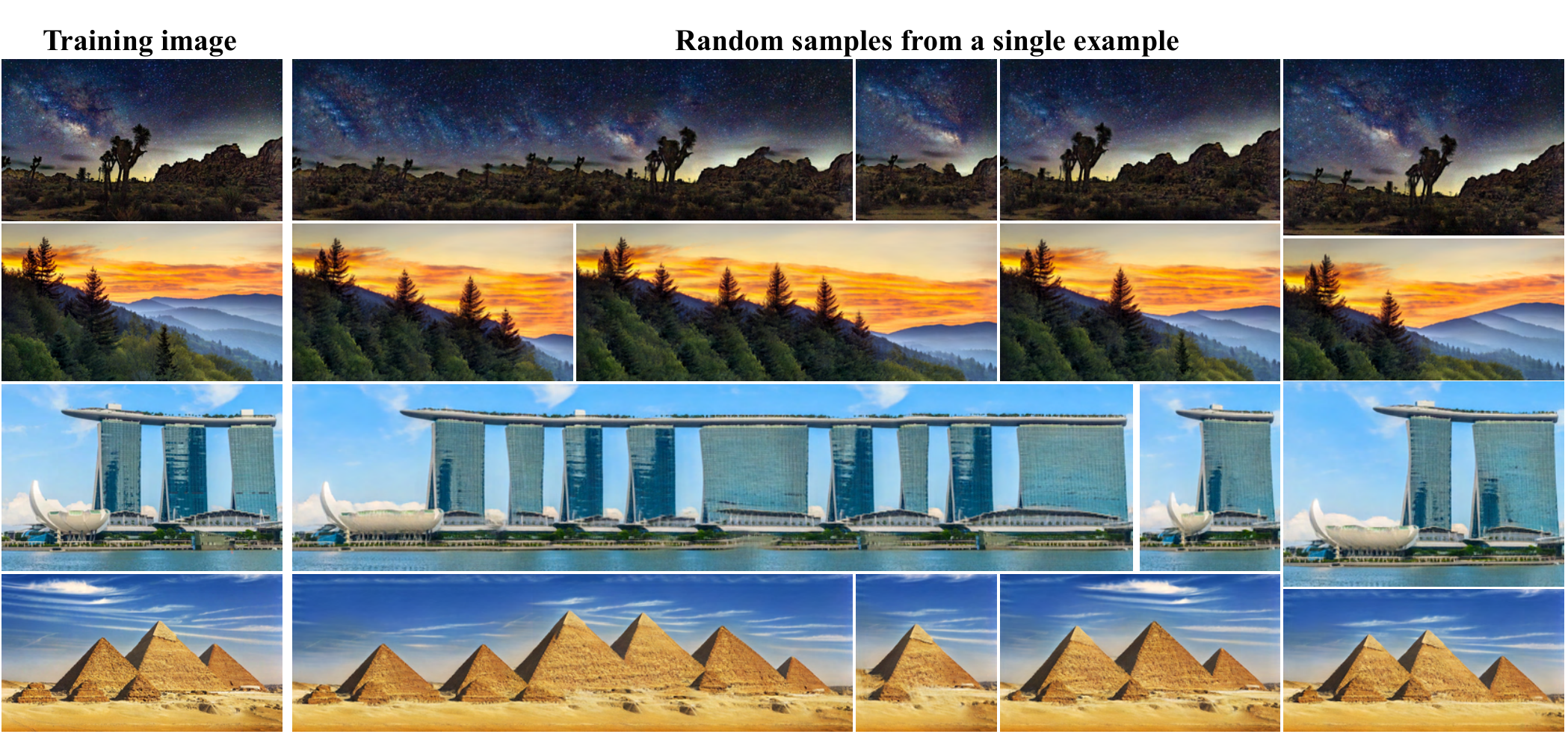}
    \caption{\textbf{Single image diffusion model.} We introduce a framework for training an unconditional denoising diffusion model (DDM) on a single image. Our single-image DDM (SinDDM) can generate novel high-quality variants of the training image at arbitrary dimensions by creating new configurations of both large objects and small-scale structures (\eg the shape of the skyline in row 1 and the angles formed by the distant mountains in row 2). SinDDM can be used for many tasks, including text-guided generation from a single image (Fig.~\ref{fig:application}).}
    \label{fig:generation}
\end{figure*}


\begin{abstract}
    Denoising diffusion models (DDMs) have led to staggering performance leaps in image generation, editing and restoration. However, existing DDMs use very large datasets for training. Here, we introduce a framework for training a DDM on a single image. Our method, which we coin SinDDM, learns the internal statistics of the training image by using a multi-scale diffusion process.
    To drive the reverse diffusion process, we use a fully-convolutional light-weight denoiser, which is conditioned on both the noise level and the scale. This architecture allows generating samples of arbitrary dimensions, in a coarse-to-fine manner. As we illustrate, SinDDM generates diverse high-quality samples, and is applicable in a wide array of tasks, including style transfer and harmonization. Furthermore, it can be easily guided by external supervision. Particularly, we demonstrate text-guided generation from a single image using a pre-trained CLIP model. Results, code and the Supplementary Material are available on the project's \href{https://matankleiner.github.io/sinddm/}{webpage}.
\end{abstract}

%% file: Intro.tex
\begin{figure*}
  \centering
   \includegraphics[width=1\linewidth]{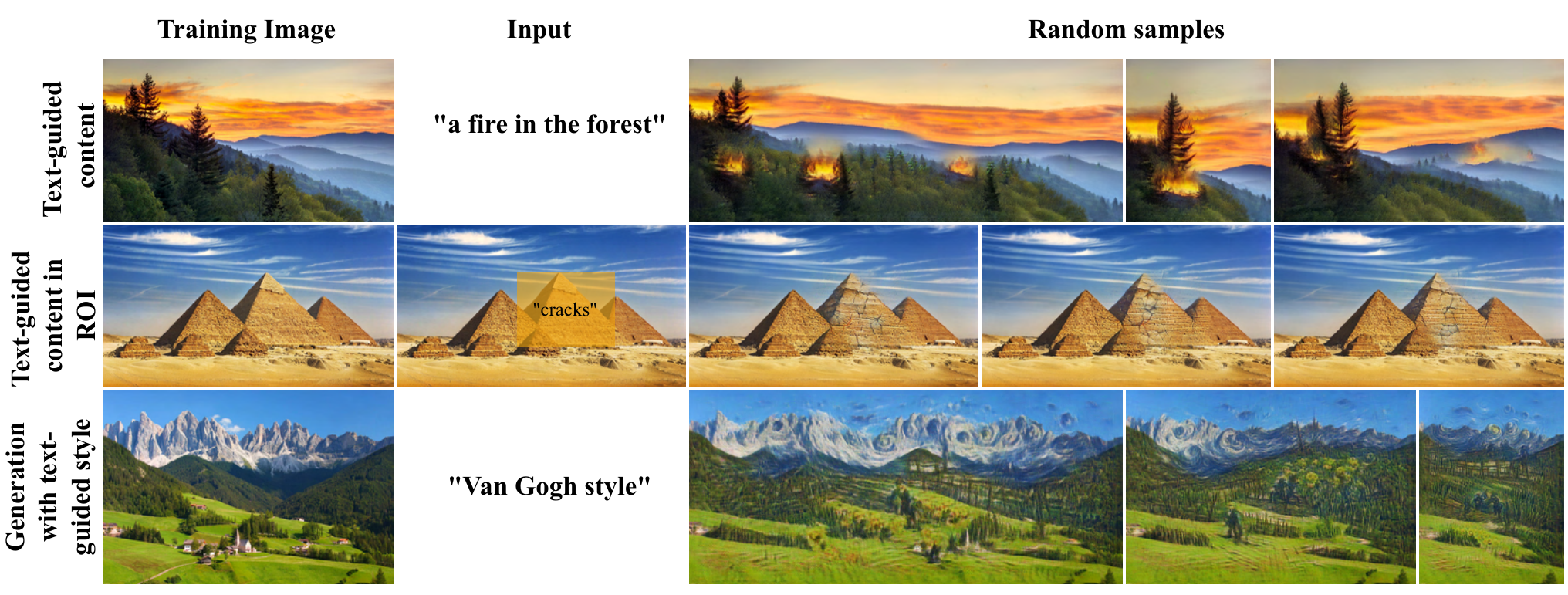}
   \caption{\textbf{Text guided generation.} SinDDM can generate images conditioned on text prompts in several different manners. We can control the contents of the generated samples across the entire image (top) or within a user-prescribed region of interest (middle). We can also control the style of the generated samples (bottom). All effects are achieved by modifying only the sampling process, without the need for any architectural changes or for training or tuning the model (see Sec.~\ref{sec:Experiments}).}
   \label{fig:application}
\end{figure*}

\section{Introduction}\label{sec:Intro}
Image synthesis and manipulation has attracted a surge of research in recent years, leading to impressive progress in \eg generative adversarial network (GAN) based methods \citep{goodfellow2020generative} and denoising diffusion models (DDMs) \citep{sohl2015deep}. State-of-the art generative models now reach high levels of photo-realism \citep{sauer2022stylegan,ho2020denoising,dhariwal2021diffusion}, can treat arbitrary image dimensions \citep{chai2022anyresolution}, can be used to solve a variety of image restoration and manipulation tasks \citep{saharia2022image,saharia2022palette,meng2021sdedit}, and can even be conditioned on complex text prompts \citep{nichol2021glide,ramesh2022hierarchical,rombach2022high,chitwan2022imagen}. However, this impressive progress has often gone hand-in-hand with the reliance on increased amounts of training data. Unfortunately, in many cases relevant training examples are scarce.

Recent works proposed to learn a generative model from a single natural image. The first \emph{unconditional} model proposed for this task was SinGAN \citep{shaham2019singan}. This model uses a pyramid of patch-GANs to learn the distribution of small patches in several image scales. Once trained on a single image, SinGAN can randomly generate similar images, as well as solve a variety of tasks, including editing, style transfer and super-resolution. Follow up works improved SinGAN's training process \citep{hinz2021improved}, extended it to other domains (\eg audio \citep{greshler2021catch}, video \citep{gur2020hierarchical}, 3D shapes \citep{wu2022learning}), and used alternative learning frameworks (energy-based models \citep{zheng2021patchwise}, nearest-neighbor patch search \citep{granot2022drop}, enforcement of deep feature statistics via test-time optimization \citep{elnekave2022generating}).

In this paper, we propose a different approach for learning a generative model from a single image. Specifically, we combine the multi-scale approach of SinGAN with the power of DDMs to devise SinDDM, a hierarchical DDM that can be trained on a single image. At the core of our method is a fully-convolutional denoiser, which we train on various scales of the image, each corrupted by various levels of noise. We take the denoiser's receptive field to be small so that it only captures the statistics of the fine details within each scale. At test time, we use this denoiser in a coarse-to-fine manner, which allows generating diverse random samples of arbitrary dimensions. As illustrated in Fig.~\ref{fig:generation}, SinDDM synthesizes high quality images while exhibiting good generalization capabilities. For example, certain small structures in the skylines in row 1 and the angles of some of the mountains in row 2 do not exist in the corresponding training images, yet they look realistic.

Similarly to existing single-image generative models, SinDDM can be used for image-manipulation tasks (see Sec.~\ref{sec:Experiments}). However, perhaps its most appealing property is that it can be easily guided by external supervision. For example, in Fig.~\ref{fig:application} we demonstrate text guidance for controlling the content and style of samples. These effects are achieved by employing a pretrained CLIP model \citep{radford2021learning}. Other guidance options are illustrated in Sec.~\ref{sec:Experiments}.

\begin{figure*}[t]
  \centering
   \includegraphics[width=1\linewidth]{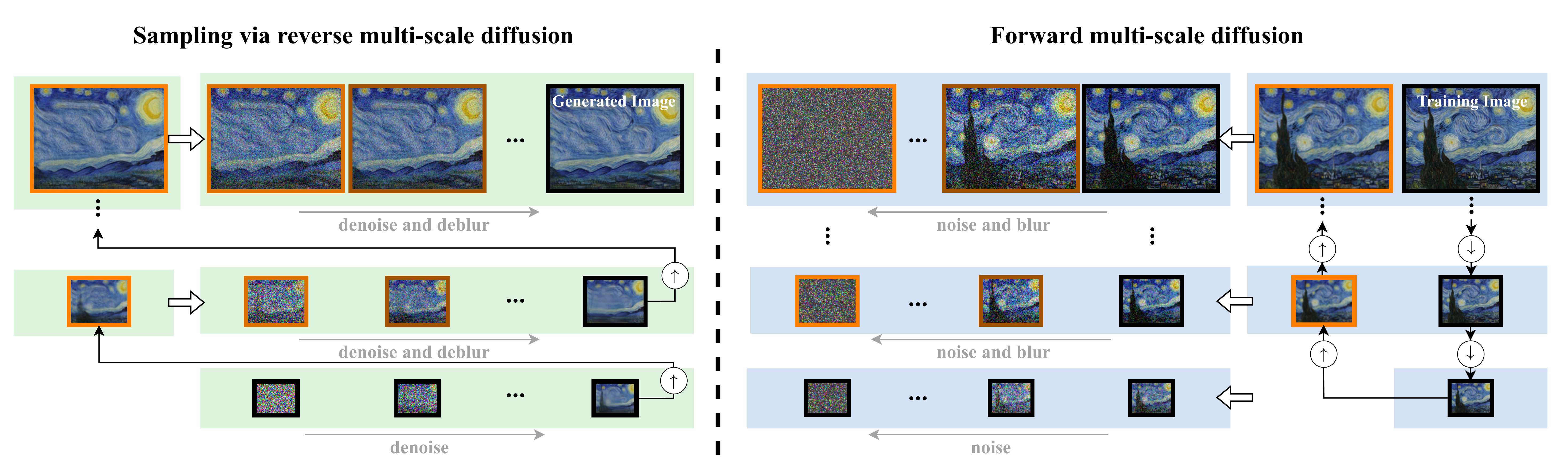} 
   \caption{\textbf{Multi-scale diffusion.} Our forward multi-scale diffusion process (right) is constructed from down-sampled versions of the training image (black frames), as well as their blurry versions (orange frames). In each scale, we construct a sequence of images that are linear combinations of the original image in that scale, its blurry version, and noise. Sampling via the reverse multi-scale diffusion (left), starts from pure noise at the coarsest scale. In each scale, our model gradually removes the noise until reaching a clean image, which is then upsampled and combined with noise to start the process again in the next scale.}
   \label{fig:sampling}
\end{figure*}

%% file: RelatedWork.tex
\section{Related Work}\label{sec:RelatedWork}

\noindent\textbf{Single image generative models\;\;\;}
Single-image generative models perform image synthesis and manipulation by capturing the internal distribution of patches or deep features within a single image. \citet{shocher2019ingan} presented a single-image conditional GAN model for the task of image retargeting. In the context of unconditional models, SinGAN \citep{shaham2019singan} is a hierarchical GAN model that can generate high quality, diverse samples based on a single training image. 
SinGAN's training process was improved by \citet{hinz2021improved}. Several works replaced SinGAN's GAN framework by other techniques for learning distributions. These include energy-based models \citep{zheng2021patchwise}, nearest-neighbor patch search \citep{granot2022drop}, and enforcement of deep-feature distributions via test-time optimization of a sliced-Wasserstein loss \cite{elnekave2022generating}. Here, we follow the hierarchical approach of SinGAN, but using denoising diffusion probabilistic models \cite{ho2020denoising}. This enables us to generate high quality images, while supporting guided image generation as in \citep{dhariwal2021diffusion}. We note that two concurrent works suggested frameworks for training a diffusion model on a single signal \citep{nikankin2022sinfusion, wang2022sindiffusion}. Those techniques differ from ours, and particularly, do not fully explore the possibilities of controlling such internal models via text-guidance.

\vspace{0.1cm}\noindent\textbf{Diffusion models\;\;\;} 
First presented by \citet{sohl2015deep}, diffusion models sample from a distribution by reversing a gradual noising (diffusion) process. This method recently achieved impressive results in image generation \citep{dhariwal2021diffusion,ho2020denoising} as well as in various other tasks, including super-resolution \citep{saharia2022image}, image-to-image translation \citep{saharia2022palette} and image editing \citep{meng2021sdedit}. These works established diffusion models as the current state-of-the-art in image generation and manipulation.

\vspace{0.1cm}\noindent\textbf{Text-guided image manipulation and generation\;\;\;} 
Text-guided image generation has recently attracted considerable interest with the emergence of models like DALL-E~2 \citep{ramesh2022hierarchical}, stable diffusion \citep{rombach2022high} and Imagen \citep{chitwan2022imagen}, which was even extended to video generation \citep{ho2022imagen}.
Besides generation, those techniques have also been found useful for image editing tasks, such as manipulating a set of user provided images using text \citep{gal2022image}. 
One popular way to guide image generation models by text is by using a pre-trained CLIP model \citep{radford2021learning}. In SinDDM we adopt this approach and combine CLIP's external knowledge with our internal model to guide the image generation process by text prompts.  
Recently, Text2Live \citep{bar2022text2live} described an approach for text-guided image editing by training on a single image. This method uses a pre-trained CLIP model to guide the generation of an edit layer that is later combined with the original image. Thus, as opposed to our goal here, Text2Live can only add details on top of the original image; it cannot change the entire scene (\eg changing object configurations) or generate images whose dimensions differ from the original image. 

%% file: Method.tex
\section{Method}\label{sec:Method}

Our goal is to train an unconditional generative model to capture the internal statistics of structures at multiple scales within a single training image. Similarly to existing DDM frameworks, we employ a diffusion process, which gradually turns the image into white Gaussian noise. However, here we do it in a hierarchical manner that combines both blur and noise. 

\subsection{Forward Multi-Scale Diffusion} As illustrated in the right pane of Fig.~\ref{fig:sampling}, we start by constructing a pyramid $\{x^{N-1}, \ldots, x^0\}$ with a scale factor of $r>0$ (black frames). Each $x^s$ is obtained by down-sampling $x$ by $r^{N-1-s}$ (so that $x^{N-1}$ is the training image $x$ itself). We also construct a blurry version of the pyramid (orange frames), $\{\tilde{x}^{N-1}, \ldots, \tilde{x}^0\}$, where $\tilde{x}^0=x^0$ and $\tilde{x}^s=(x^{s-1})\!\!\uparrow^r$ for every $s\geq 1$. We use bicubic interpolation for both the upsampling and downsampling operations. We use those two pyramids to define a multi-scale diffusion process over $(s,t)\in\{0,\ldots,N-1\}\times\{0,\ldots,T\}$ as
\begin{align}    
    x_{t}^{s} = \sqrt{\bar{\alpha}_{t}}\left(\gamma^s_t \tilde{x}^s + (1-\gamma^s_t)x^s\right) +\sqrt{1-\bar{\alpha}_{t}}\,\epsilon\,,
\end{align}
where $\epsilon\sim\mathcal{N}(0,\boldsymbol{I})$. As $t$ grows from $0$ to $T$, $\gamma^s_t$ increases monotonically from $0$ to $1$ while $\bar{\alpha}_{t}$ decreases monotonically from $1$ to~$0$ (see Appendix \ref{appendix:diff_process} for details). Therefore, as $t$ increases, $x_{t}^{s}$ becomes both noisier and blurrier.
The reason for using a blurry version of the image in each scale, is associated with the sampling process, as we explain next.


\begin{figure}[t]
  \centering
   \includegraphics[width=0.7\linewidth]{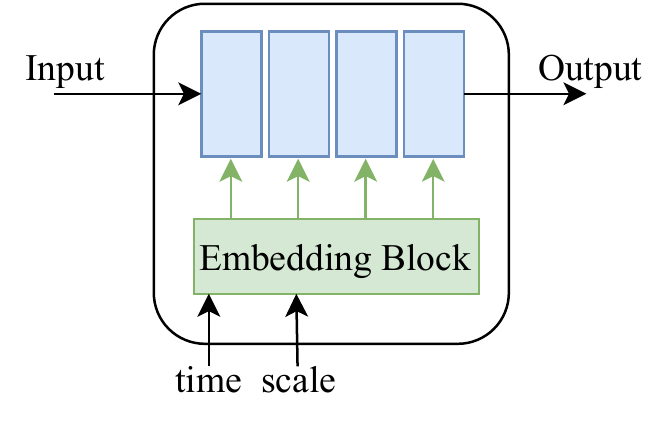}
   \caption{\textbf{SinDDM architecture.} We use a fully-convolutional model with four blocks, having a total receptive field of $35\times35$. The model is conditioned on both the timestep $t$ and the scale $s$.}
   \label{fig:arch}
\end{figure}

\begin{algorithm}[t]
\caption{SinDDM Training}\label{alg:training}
\begin{algorithmic}[1]
\REPEAT
\STATE $s \sim \text{Uniform}(\{0,...,N-1\})$
\STATE $t \sim\text{Uniform}(\{0,...,T\})$
\STATE $\epsilon\sim\mathcal{N}(0,\boldsymbol{I})$
\IF {$s=0$}
    \STATE $x_{t}^{s,\text{mix}} = x^s$    
\ELSE
    \STATE $x_{t}^{s,\text{mix}} = \gamma^s_t \,x^{s-1}\uparrow^r + (1 - \gamma^s_t) x^s$
\ENDIF
\STATE Update model $\epsilon_{\theta}$ by taking gradient descent step on
\STATE \ \ \ \ \ \ \ \ \  $\nabla_{\theta} \left\|\epsilon-\epsilon_{\theta}(\sqrt{\bar{\alpha}_{t}} x_{t}^{s,\text{mix}}+\sqrt{1-\bar{\alpha}_{t}}\epsilon,t,s)\right\|_{1}$
\UNTIL{converged}
\end{algorithmic}
\end{algorithm}

\begin{algorithm}[t]
\caption{SinDDM Sampling}\label{alg:sampling}
\begin{algorithmic}[1]
\FOR{$s = 0,\ldots,N-1$}
\IF{$s=0$}
\STATE $x^{0}_{T[0]}\sim \mathcal{N}(0, \boldsymbol{I})$
\ENDIF
\FOR{$t=T[s],\ldots,1$}
 \STATE $x^{s,\text{mix}}_{t}=\frac{x_t^s-\sqrt{1-\bar{\alpha}_t}\,\epsilon_\theta(x^s_t,t,s)}{\sqrt{\bar{\alpha}_t}}$
\STATE $\hat{x}^{s}_{0}=\frac{x_t^{s,\text{mix}}-\gamma_t^s\tilde{x}^s}{1-\gamma^s_t}$


\STATE $x_{t-1}^{s,\text{mix}}=\gamma^s_{t-1} \tilde{x}^s  + (1-\gamma^s_{t-1})\hat{x}^s_0$

\STATE $z\sim \mathcal{N}(0, \boldsymbol{I})$
\STATE $x_{t-1}^s = \sqrt{\bar{\alpha}_{t-1}}x^{s,\text{mix}}_{t-1}$
\STATE $\quad\quad+ \sqrt{1-\bar{\alpha}_{t-1}-(\sigma_t^s)^2} \,\frac{x^s_t-\sqrt{\bar{\alpha}_t}x^{s,\text{mix}}_{t}}{\sqrt{1-\bar{\alpha}_{t}}} + \sigma_t^s z$

\ENDFOR
\IF{$s < N-1$}
\STATE $\tilde{x}^{s+1}=\hat{x}^{s}_{0}\uparrow^r$
\STATE $z\sim \mathcal{N}(0, \boldsymbol{I})$
\STATE $x^{s+1}_{T[s+1]}=\sqrt{\bar{\alpha}_{T[s+1]}}\tilde{x}^{s+1} +\sqrt{1-\bar{\alpha}_{T[s+1]}}z$ 
\ENDIF
\ENDFOR
\end{algorithmic}
\end{algorithm}

\subsection{Reverse Multi-Scale Diffusion}
To sample an image, we attempt to reverse the diffusion process, as shown in the left pane of Fig.~\ref{fig:sampling}. Specifically, we start at scale $s=0$, where we follow the standard DDM approach (starting with random noise at $t=T$ and gradually removing noise until a clean sample is obtained at $t=0$). We then upsample the generated image to scale $s=1$, combine it with noise again, and run a reverse diffusion process to form a sample at this scale. The process is repeated until reaching the finest scale, $s=N-1$. 

\begin{figure*}[h]
  \centering
  \includegraphics[width=\textwidth]{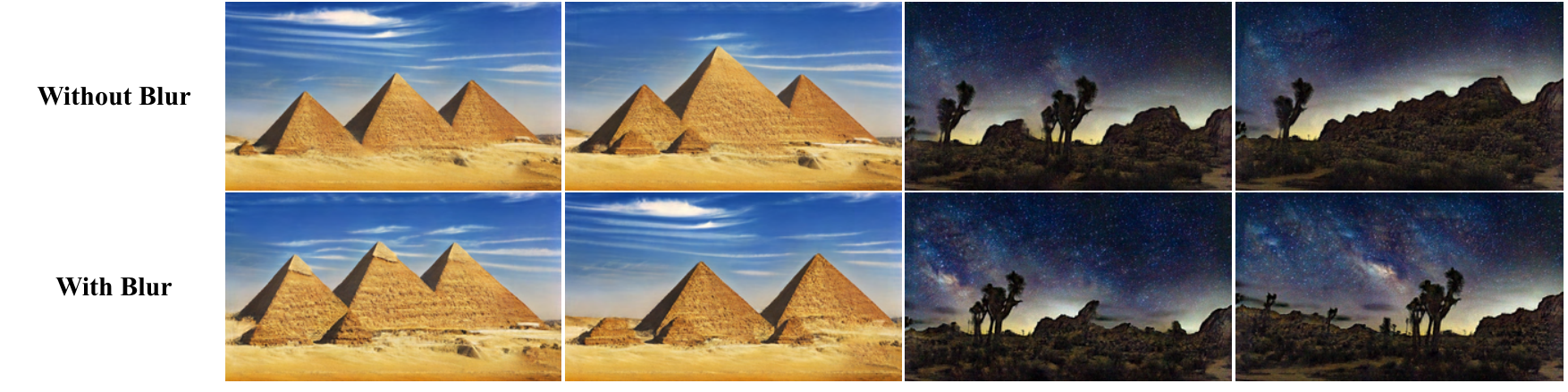}
  \caption{\textbf{Training with and without blur.} For each training image, we compare samples from two models trained on that image, one trained without blur and one trained with blur. The images sampled from the model that was trained without blur lack fine details such as the pyramids' texture and the stars in the night sky.}
  \label{fig:blur_comp}
\end{figure*}

Note that since we upsample the image between scales, we naturally add blur. This implies that our model needs to remove not only noise, but also blur. This is the reason that during the forward process, we also gradually blur the image in addition to adding noise (for every scale $s>0$). This forces the model to learn to remove both noise and blur from the initial image. The importance of adding blur is illustrated in Fig.~\ref{fig:blur_comp}. 

The reverse diffusion process is driven by a single fully convolutional model, which is trained to predict $x_{0}^s$ based on $x_t^s$ (in practice it predicts the noise $\epsilon$ from which we extract a prediction of $x_{0}^s$). The training procedure is shown in Alg.~\ref{alg:training}. For sampling, we adopt the DDIM formulation \citep{song2020denoising}, as detailed in Alg.~\ref{alg:sampling}, where for scale $s=0$ we use the noise variance of DDPM \citep{ho2020denoising} and for $s>0$ we use $\sigma_t^s=0$ except when applying text-guidance, in which case we also use the DDPM scheduler. Note that for $s=0$, $\gamma_t^s=0$ since there is no blur to remove. More details about the $\gamma_t^s$ schedule in regular sampling and in text-guided sampling are provided in appendices~\ref{appendix:diff_process} and~\ref{appendix:text_gen}.

As shown in Fig.~\ref{fig:arch}, our model is conditioned on both the timestep $t$ and the scale~$s$.
We found this to improve generation quality and training time compared to a separate diffusion model for each scale. Our model comprises 4 convolutional blocks, with a total receptive field of $35\times 35$. The number of scales is chosen such that the area covered by the receptive field is as close as possible to $40\%$ of the area of the entire image at scale 0. In most of our experiments, this rule led to 4 or 5 scales. The small receptive field forces the model to learn the statistics of small structures and prevents memorization of the entire image.
For every scale $s>0$, we start the reverse diffusion at timestep $T[s]\leq T$, which we set such that $(1-\bar{\alpha}_{T[s]})/\bar{\alpha}_{T[s]}$ is proportional to the MSE between $x^s$ and $\tilde{x}^s$. This ensures that the amount of noise added to the upsampled image from the previous scale is proportional to the amount of missing details at that scale (see derivation in App.~\ref{appendix:diff_process}). For $s=0$, we start at $T[0]=T$.

As opposed to external DDMs, our model uses only convolutions and GeLU nonlinearities, without any self-attention or downsampling/upsampling operations. The timestep $t$ and scale $s$ are injected to the model using a joint embedding, similarly to the one used to inject only $t$ in \citep{ho2020denoising} (see App.~\ref{appendix:Arch}). The model has a total of $1.1\times10^6$ parameters and its training on a $250\times200$ image takes around 7 hours on an A6000 GPU. Sampling of a single image takes a few seconds. In each training iteration we sample a batch of noisy images from the same randomly chosen scale $s$ but from several randomly chosen timesteps $t$. We train the model for 120,000 steps using the Adam optimizer with its default parameters (see App.~\ref{appendix:Training} for further details).

\begin{figure*}
  \centering
   \includegraphics[width=0.95\linewidth]{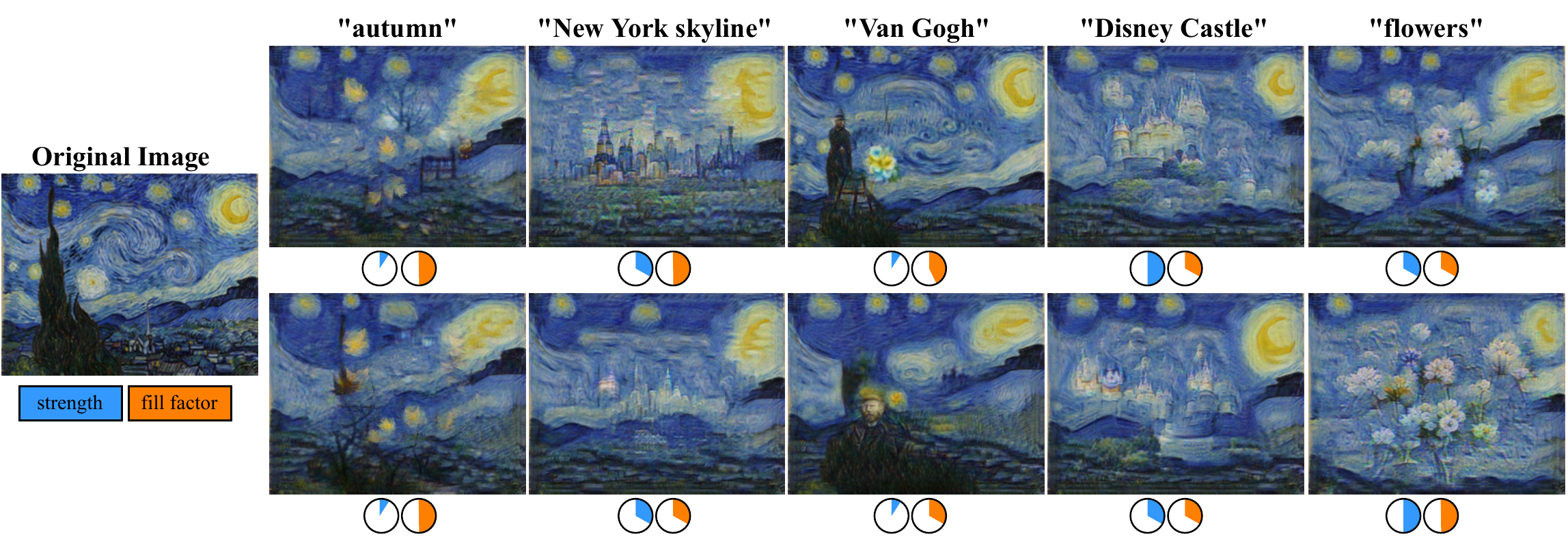}
   \caption{\textbf{Image generation guided by text.}  SinDDM can generate diverse samples guided by a text prompt. The strength of the effect is controlled by the strength parameter $\eta$ (blue), while the spatial extent of the affected regions is controlled by the fill factor $f$ (orange).}
   \label{fig:clip}
\end{figure*}

\subsection{Guided Generation}
To guide the generation by a user-provided loss, we 
follow the general approach of \citet{dhariwal2021diffusion}, where the gradient of the loss is added to the predicted clean image in each diffusion step. Here we describe two ways to guide SinDDM generations, one by choosing a region of interest (ROI) in the original image and its desired location in the generated image and one by providing a text prompt.  

\paragraph{Generation guided by image ROIs}
In image-guided generation, the user chooses regions from the training image and selects where they should appear in the generated image. The rest of the image is generated randomly, but coherently with the constrained regions (see Fig.~\ref{fig:roi_guidance}). To achieve this effect, we use a simple $L^2$ loss. Specifically, let $x_{\text{target}}^s$ be an image containing the desired contents within the target ROIs and let $m^s$ be a binary mask indicating the ROIs, both down-sampled to scale $s$. Then we define our ROI guidance loss to be $\mathcal{L}_{\text{ROI}}=\|m^s\odot(\hat{x}^s_0- x^s_{\text{target}})\|^2$. Taking a gradient step on this loss boils down to replacing the current estimate of the clean image, $\hat{x}^s_0$, by a linear interpolation between $\hat{x}^s_0$ and $x^s_{\text{target}}$. Namely,
\begin{equation} \label{eq:roi_samlping}
    \hat{x}^s_0\leftarrow m^s\odot((1-\eta)\hat{x}^s_0+\eta x^s_{\text{target}})+(1-m^s)\odot \hat{x}^s_0,
\end{equation}
where the step size $\eta$ determines the strength of the effect. We use this guidance in all scales except for the finest one.

\paragraph{Text guided style}
For text-guidance, we use a pre-trained CLIP model. 
Specifically, in each diffusion step we use CLIP to measure the discrepancy between our current generated image,~$\hat{x}_{0}^s$, and the user's text prompt. We do this by augmenting both the image and the text prompt, as described in \citep{bar2022text2live} (with some additional text augmentations described in App.~\ref{appendix:data_aug}), and feeding all augmentations into CLIP's image encoder and text encoder. Our loss, $\mathcal{L}_{\text{CLIP}}$, is the average cosine distance between the augmented text embeddings and the augmented image embeddings. We update $\hat{x}_{0}^s$ based on the gradient of $\mathcal{L}_{\text{CLIP}}$. 
At the finest scale $s=N-1$, we finish the generation process with three diffusion steps without CLIP guidance. Those steps smoothly blend the objects created by CLIP into the generated image. For style guidance, we provide a text prompt of the form ``X style'' (\eg ``Van Gogh style'') and apply CLIP guidance only at the finest scale. To \emph{control the style of random samples}, all pyramid levels before that scale generate a random sample as usual and are thus responsible for the global structure of the final sample. To \emph{control the style of the training image} itself we inject that image directly to the finest scale, so that the modifications imposed by our denoiser and by the CLIP guidance only affect the fine textures. This leads to a style-transfer effect, but where the style is dictated by a text prompt rather than by an example style image (see Fig.~\ref{fig:starting_from_the_image} in the appendix). 


\paragraph{Text guided contents}
To control contents using text, we use the same approach as above, but apply the guidance at all scales except $s=0$. We also constrain the spatial extent of the affected regions by zeroing out all gradients outside a mask $m^s$. This mask is calculated in the first step CLIP is applied, and is kept fixed for all remaining timesteps and scales (it is upsampled when going up the scales of the pyramid). The mask is taken to be the set of pixels containing the top $f$-quantile of the values of $\nabla_{\hat{x}_{0}^s}\mathcal{L}_{\text{CLIP}}$, where $f\in[0,1]$ is a user-prescribed \emph{fill factor}. 
We use an adaptive step size strategy, where we update $\hat{x}_{0}^s$ as
\begin{equation}
  \hat{x}_{0}^{s} \leftarrow \eta\; \delta \; m^s \odot \nabla\mathcal{L}_{\text{CLIP}}  + 
  (1-m^s)\odot \hat{x}_{0}^{s}.
  \label{eq:clip}
\end{equation}
Here $\delta=\|\hat{x}_{0}^{s} \odot m\|/\| \nabla\mathcal{L}_{\text{CLIP}} \odot m \|$ and
 $\eta \in [0,1]$ is a \emph{strength} parameter that controls the intensity of the CLIP guidance. We also use a momentum on top of this update scheme (see App.~\ref{appendix:algo}). We let the user choose both the fill factor $f$ and the strength $\eta$ to achieve the desired effect. Their influence is demonstrated in Fig.~\ref{fig:clip}.

%% file: Experiments.tex
\section{Experiments}\label{sec:Experiments}
We trained SinDDM on images of different styles, including urban and nature scenery as well as art paintings. We now illustrate its utility in a variety of tasks.


\begin{table*}[t]
\centering
\caption{Quantitative evaluation for unconditional generation. Best and second best results are marked in blue, with and without boldface fonts respectively.} 
  \begin{tabular}{c c c c c c} 
    \toprule
    Type & Metric & SinGAN & ConSinGAN & GPNN & SinDDM\\
    \midrule
    \multirow{2}{*}{Diversity} & Pixel Div. $\uparrow$ & \textcolor{blue}{0.28$\pm$0.15} & 0.25$\pm$0.2 & 0.25$\pm$0.2 & \textbf{\textcolor{blue}{0.32$\pm$0.13}}\\ 
     & LPIPS Div. $\uparrow$ & \textcolor{blue}{0.18$\pm$0.07} & 0.15$\pm$0.07 & 0.1$\pm$0.07 & \textbf{\textcolor{blue}{0.21$\pm$0.08}} \\
    \hline
    \multirow{3}{*}{No reference IQA} & NIQE $\downarrow$ & 7.3$\pm$1.5 & \textbf{\textcolor{blue}{6.4$\pm$0.9}} & 7.7$\pm$2.2 & \textcolor{blue}{7.1$\pm$1.9}\\ 
     & NIMA $\uparrow$ & \textcolor{blue}{5.6$\pm$0.5} & 5.5$\pm$0.6 & \textcolor{blue}{5.6$\pm$0.7} & \textbf{\textcolor{blue}{5.8$\pm$0.6}} \\
     & MUSIQ $\uparrow$ & 43$\pm$9.1 & 45.6$\pm$9 & \textbf{\textcolor{blue}{52.8$\pm$10.9}} & \textcolor{blue}{48$\pm$9.8}  \\ 
    \hline
    Patch Distribution & SIFID $\downarrow$ & 0.15$\pm$0.05 & \textcolor{blue}{0.09$\pm$0.05} & \textbf{\textcolor{blue}{0.05$\pm$0.04}} & 0.34$\pm$0.3\\ 
    \bottomrule
  \end{tabular}
  \label{tab:quantitative}
\end{table*}

\begin{figure*}
  \centering
   \includegraphics[width=1\linewidth]{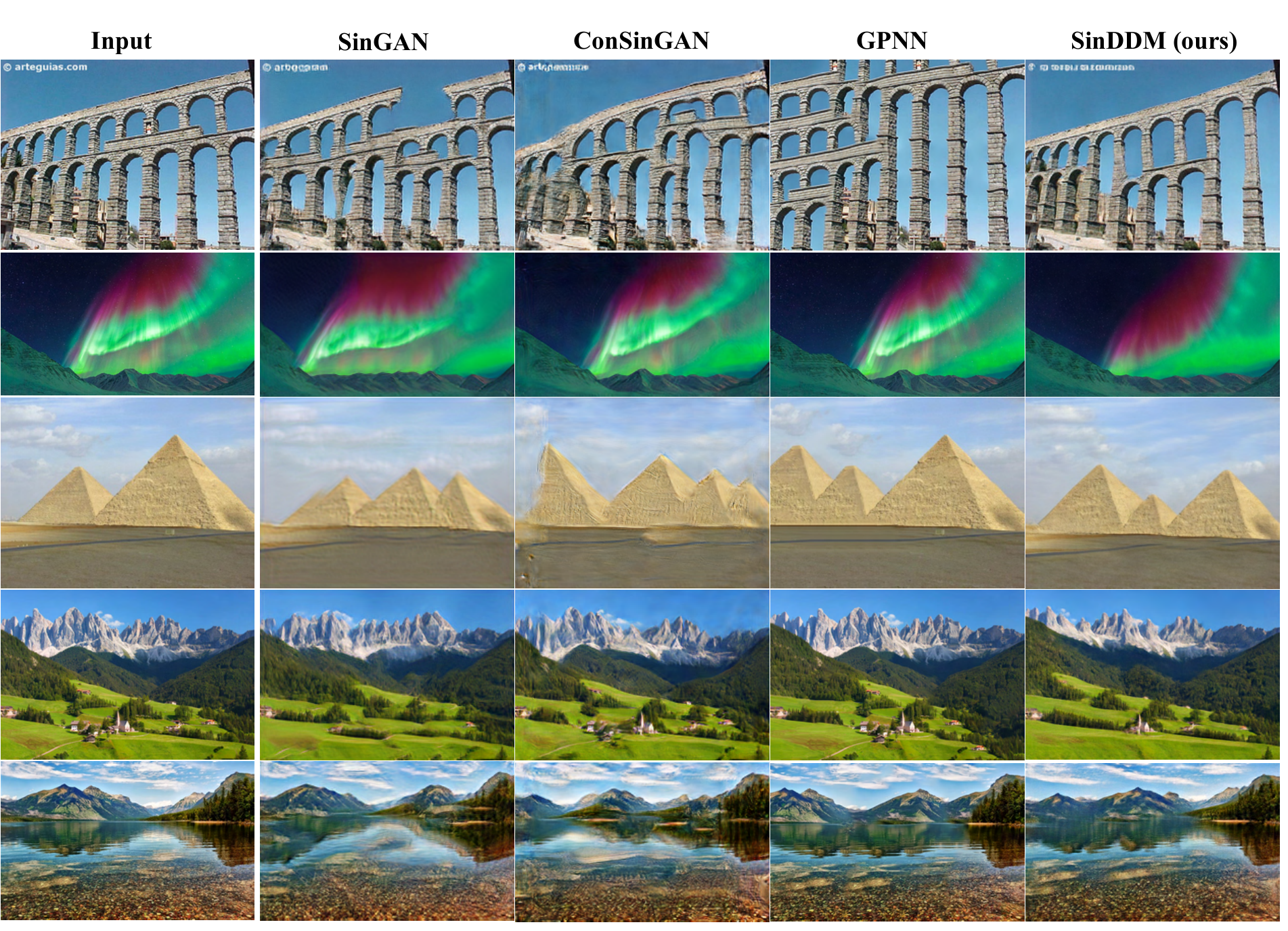}
   \caption{\textbf{Unconditional image generation comparisons.} We qualitatively compare our model to other single image generative models on unconditional image generation. As can be seen, our results are at least on par with the other models in terms of quality and generalization.}
   \label{fig:generation_comp}
\end{figure*}

\paragraph{Unconditional image generation}
As illustrated in Figs.~\ref{fig:generation}, \ref{fig:generation_comp}, \ref{fig:gen_sup} and \ref{fig:gen_comp_sup}, SinDDM is able to generate diverse, high quality samples of arbitrary dimensions. Close inspection reveals that SinDDM often generalizes beyond the structures appearing in the training image. For example, in Fig.~\ref{fig:generation}, 2nd row, the angles of many of the mountains in the leftmost sample do not appear in the training image. Table~\ref{tab:quantitative} reports a quantitative comparison to other single image generative models on all 12 images appearing in this paper (see App.~\ref{appendix:comp_sampling} for more comparisons). Each measure in the table is computed over 50 samples per training image (we report mean and standard deviation). 
As can be seen, the diversity of our generated samples (both pixel standard-deviation and average LPIPS distance between pairs of samples) is higher than the competing methods. At the same time, our samples have comparable quality to those of the competing methods, as ranked by the no-reference image quality  measures NIQE \citep{mittal2012making}, NIMA \citep{talebi2018nima} and MUSIQ \citep{ke2021musiq}.  However, the single-image FID (SIFID) \citep{shaham2019singan} achieved by SinDDM is higher than the competing methods. This is indicative of the fact that SinDDM generalizes beyond the structures in the training image, so that the internal deep-feature distributions are not preserved. Yet, as we show next, this does not prevent from obtaining highly satisfactory results in a wide range of image manipulation tasks.

\begin{figure*}
  \centering
   \includegraphics[width=1\linewidth]{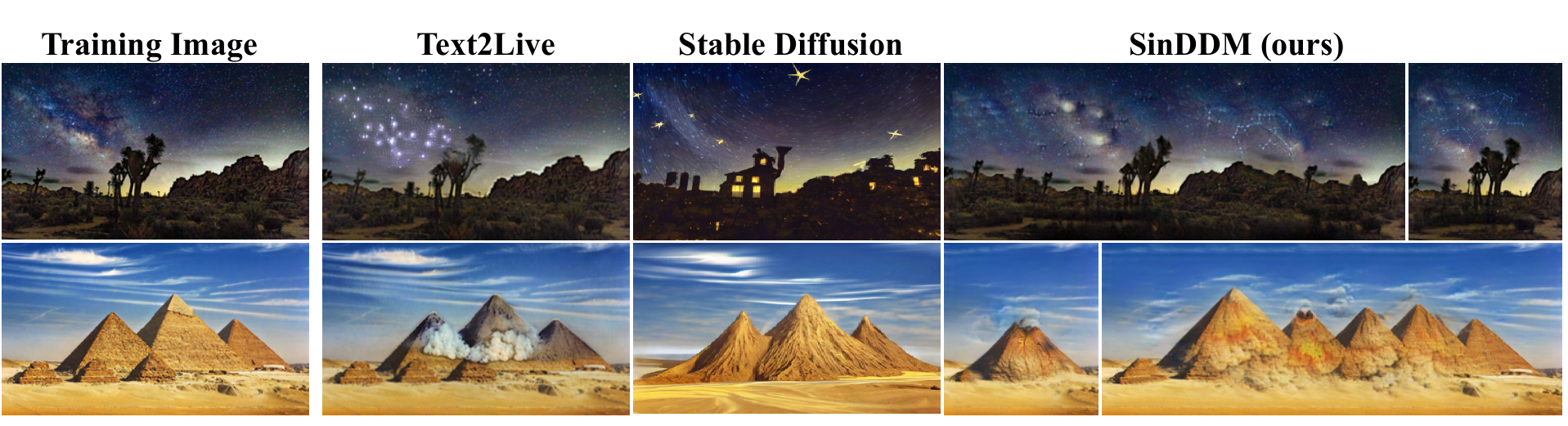}
   \caption{\textbf{Image generation and editing guided by text.} We compare SinDDM to Text2Live and stable diffusion (using the approach of SDEdit). Unlike these methods, SinDDM is not constrained to the aspect ratio or scene arrangement of the training image. We used the text prompts ``\emph{stars constellations in the night sky}'' and ``\emph{volcano eruption}'' for the 1st and 2nd rows, respectively. Text2Live requires four different text prompt as inputs. For the pyramid image, we supplied it with the additional texts ``volcano erupt from the pyramids in the desert'' to describe the full edited image, ``pyramids in the desert'' to describe the input image and ``the pyramids'' to describe the ROI in the input image (see App.~\ref{appendix:comp_content} for the text prompts we used for the night sky image). For stable diffusion we tried many strength values and chose the best result (see App.~\ref{appendix:comp_content} for other strengths).}
   \label{fig:clip_comp}
\end{figure*}

\begin{figure*}
  \centering
   \includegraphics[width=1\linewidth]{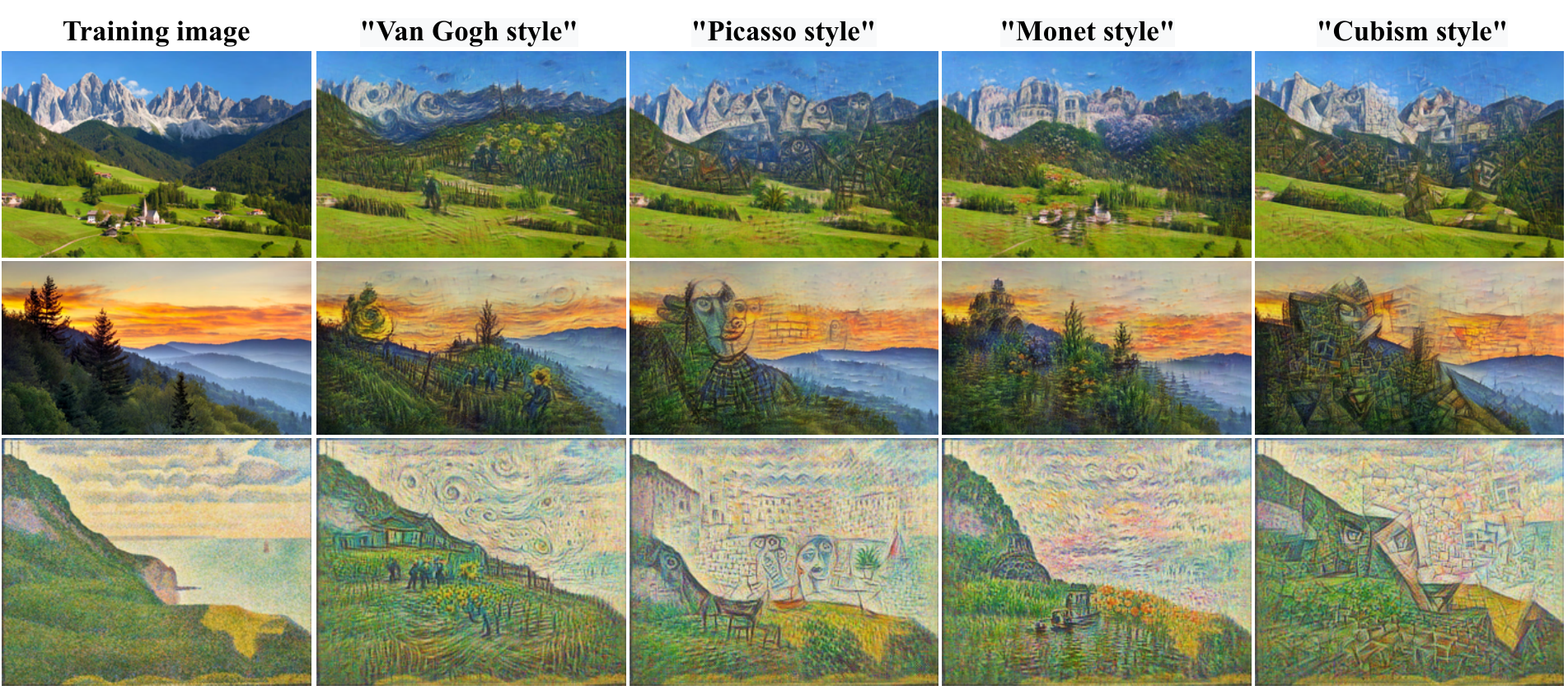}
   \caption{\textbf{Generation with text guided style.} SinDDM can generate samples in a prescribed style using CLIP guidance at the finest scale.}
   \label{fig:style_text}
\end{figure*}
\paragraph{Generation with text guided contents}
Figures~\ref{fig:application}, \ref{fig:clip}, \ref{fig:gen_content_sup} present text guided content generation examples. As can be seen, our approach allows obtaining quite significant effects, while also remaining loyal to the internal statistics of the training image. 
In Figs.~\ref{fig:application} and \ref{fig:clip_roi} we illustrate editing of local regions via text. In this setting, the user chooses a ROI and a corresponding text prompt. These are used as inputs to CLIP's image and text encoders, and the gradients of the CLIP loss are used to modify only the ROI. 
In Figs.~\ref{fig:clip_comp} and \ref{fig:clip_comp_sup1}-\ref{fig:clip_comp_sup3} we compare our text-guided content generation method to Text2Live \citep{bar2022text2live} and to Stable Diffusion \citep{rombach2022high}. Text2Live is an image editing method that can operate on any image (or video) using a text prompt. It does so by synthesizing an edit layer on top of the original image. The edit is guided by four different text prompts that describe the input image, the edit layer, the edited image and the ROI. This method cannot move objects, modify scene arrangement, or generate images of different aspect ratios. Our model is guided only by one text prompt that describes the desired result and can generate diverse samples of arbitrary dimensions. As for Stable Diffusion, we use the ``image-to-image'' option implemented in their source code. In this setting, the image is embedded into a latent space and injected with noise (controlled by a strength parameter). The denoising process is guided by the user's text prompt, similarly to the framework described in SDEdit \citep{meng2021sdedit} 
(see App.~\ref{appendix:comp_content}). 


\paragraph{Generation with text guided style}
Figures~\ref{fig:application}, \ref{fig:style_text}, \ref{fig:style_a_d}-\ref{fig:painting} present examples of image generation with a text-guided style. Here, the guidance generates not only the textures and brush strokes typical of the desired style, it also generates fine semantic details that are commonly seen in paintings of this style (\eg typical scenery, sunflowers in ``Van Gogh style''). Figure~\ref{fig:starting_from_the_image} shows text-guided style transfer.

\begin{figure*}[t]
    \centering
    \includegraphics[width=1\linewidth]{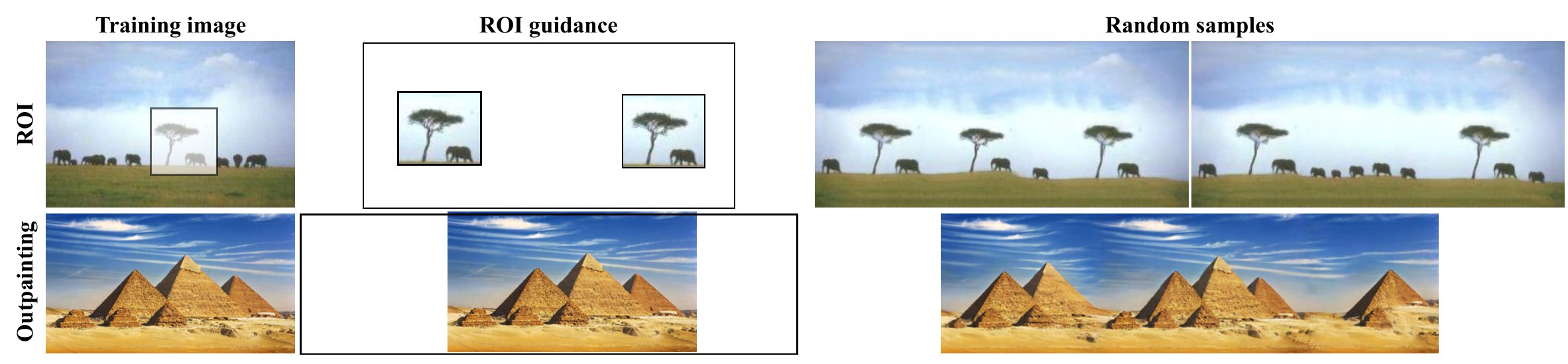}
    \caption{\textbf{Image guided generation in ROIs.} Our model is able to generate images with user-prescribed contents within several ROIs. The rest of the image is generated randomly but coherently around those constraints.The first row exemplifies enforcing two identical ROIs. The second row demonstrates selecting the entire image as the ROI and a wider target image, resulting in an outpainting effect.}
    \label{fig:roi_guidance}
\end{figure*}

\paragraph{Generation guided by image ROIs}
Figures~\ref{fig:roi_guidance} and \ref{fig:ROI_sup} show examples for generation guided by image ROIs. Here, the goal is to generate samples while forcing one or more ROIs to contain pre-determined content. 
As we illustrate, this particularly allows to perform outpainting. This is done by letting the ROI be the entire training image and generating samples with a larger aspect ratio (\eg twice as large horizontally). SinDDM generates diverse contents outside the constrained ROIs that coherently stitch with the constrained regions.

\begin{figure}[t]
  \centering
   \includegraphics[width=\linewidth]{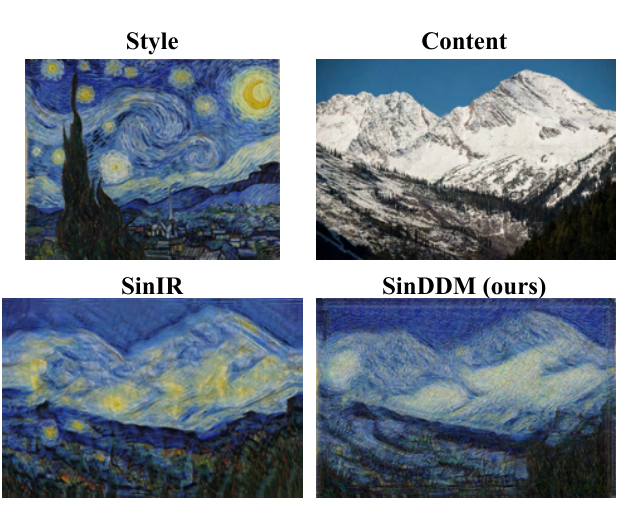}
   \caption{\textbf{Style transfer.} SinDDM can transfer the style of the training image to a content image, while preserving the global structure of the content image. }
   \label{fig:style_transfer}
\end{figure}

\paragraph{Style transfer}
Similarly to SinGAN, SinDDM can also be used for image manipulation tasks, by relying on the fact that it can only sample images with the internal statistics of the training image. Particularly, to perform style transfer, we train our model on the style image and inject a downsampled version of the content image into some scale $s \leq N-1$ and timestep $t\leq T$ (by adding noise with the appropriate intensity). We then run the reverse multi-scale diffusion process to obtain a sample.  At the injection scale, we use $\gamma^s_t=0$ for all $t$ and match the histogram of the content image to that of the style image. As can be seen in Figs.~\ref{fig:style_transfer} and \ref{fig:style}, this leads to samples with the global structure of the content image and the textures of the style image. We show a qualitative comparison with SinIR \citep{yoo2021sinir}, a state-of-the-art internal method for image manipulation. 



\begin{figure}[t!]
  \centering
   \includegraphics[width=\linewidth, trim=0cm 0cm 0.3cm 0cm, clip]{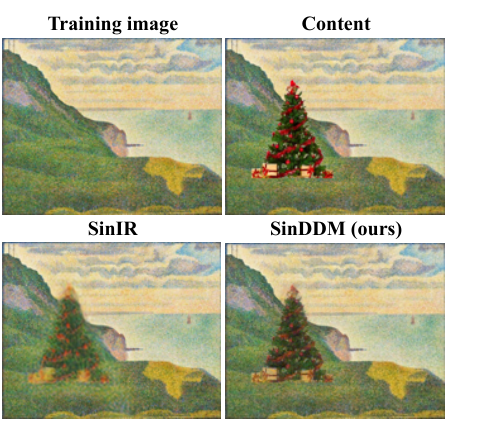}
   \caption{\textbf{Harmonization.} 
   Injecting an image with a naively pasted object into an intermediate scale and timestep, matches the object's appearance to the training image.}
   \label{fig:harmonization}
\end{figure}

\paragraph{Harmonization}
Here, the goal is to realistically blend a pasted object into a background image. To achieve this effect, we train SinDDM on the background image and inject a downsampled version of the naively pasted composite into some scale $s$ and timestep $t$ (with $\gamma^s_t=0$ at the injection scale). 
As can be seen in Fig.~\ref{fig:harmonization} and \ref{fig:harmo}, SinDDM blends the pasted object into the background, while tailoring its texture to match the background. Here, our result is less blurry than SinIR's. 



%% file: Conclusion.tex
\section{Conclusion}\label{sec:Conclusion}
We presented SinDDM, a single image generative model that combines the power and flexibility of DDMs with the multi-scale structure of SinGAN. SinDDM can be easily guided by external sources. Particularly, we demonstrated text-guided image generation, where we controlled the contents and style of the samples. 
A limitation of our method is that it is often less confined to the internal statistics of the training image than other single image generative techniques. While this can be advantageous in tasks like style transfer (see the colors in Fig.~\ref{fig:style_transfer}), in unconditional image generation, this can lead to over- or under-representation of objects in the image (see App.~\ref{appendix:FutureWork}). 

\paragraph{Acknowledgements} This research was partially supported by the Israel Science
Foundation (grant no.~2318/22) and by the Ollendorff Miverva Center, ECE faculty, Technion.

%% file: Acknowledgments.tex

%% file: Appendix.tex
\section{Additional Examples}\label{appendix:Examples}

\begin{figure}[h]
  \centering
  \includegraphics[width=\textwidth]{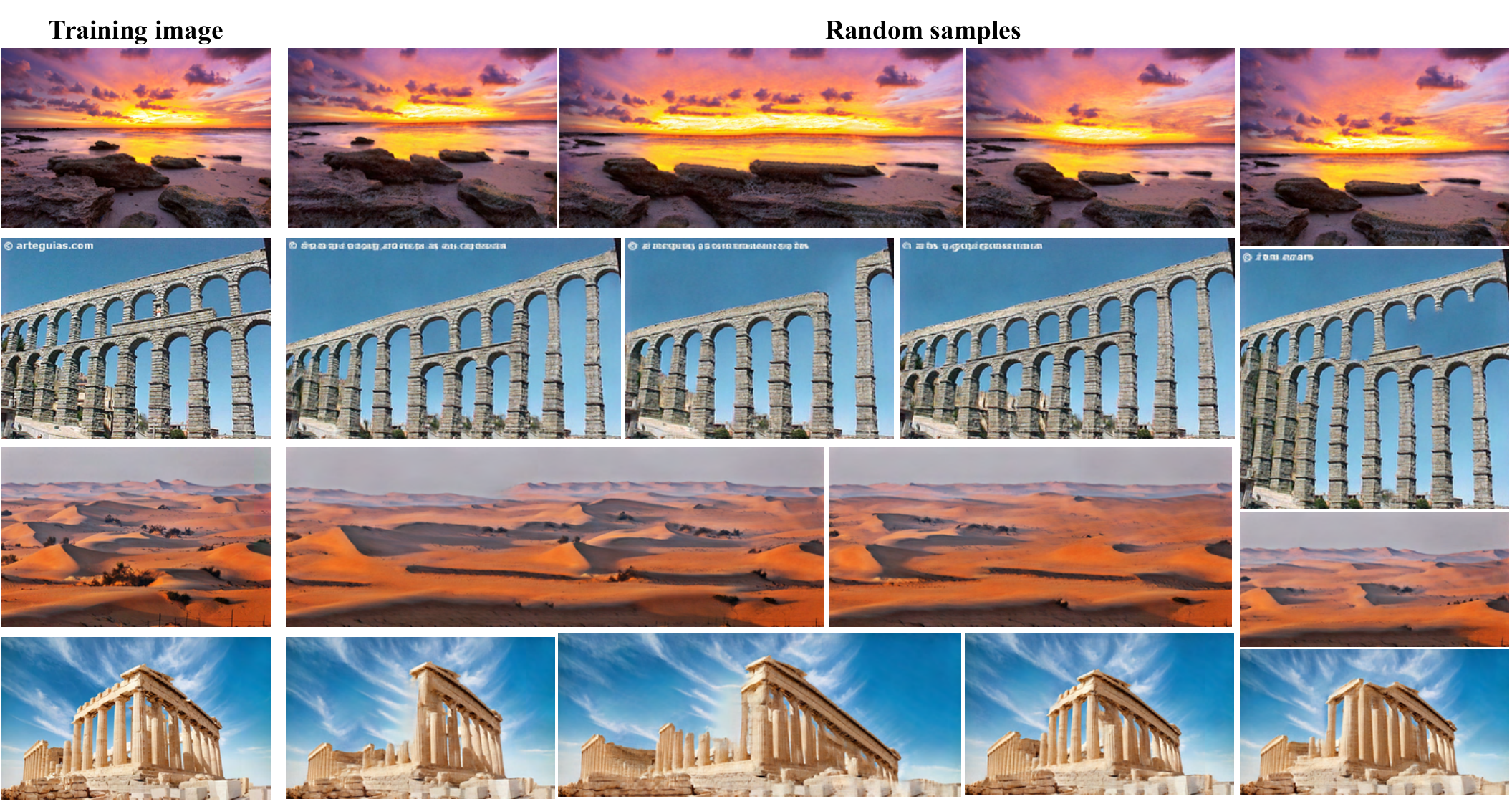}
  \caption{\textbf{Unconditional image generation.}}
  \label{fig:gen_sup}
\end{figure}

\begin{figure}[h]
  \centering
  \includegraphics[width=\textwidth]{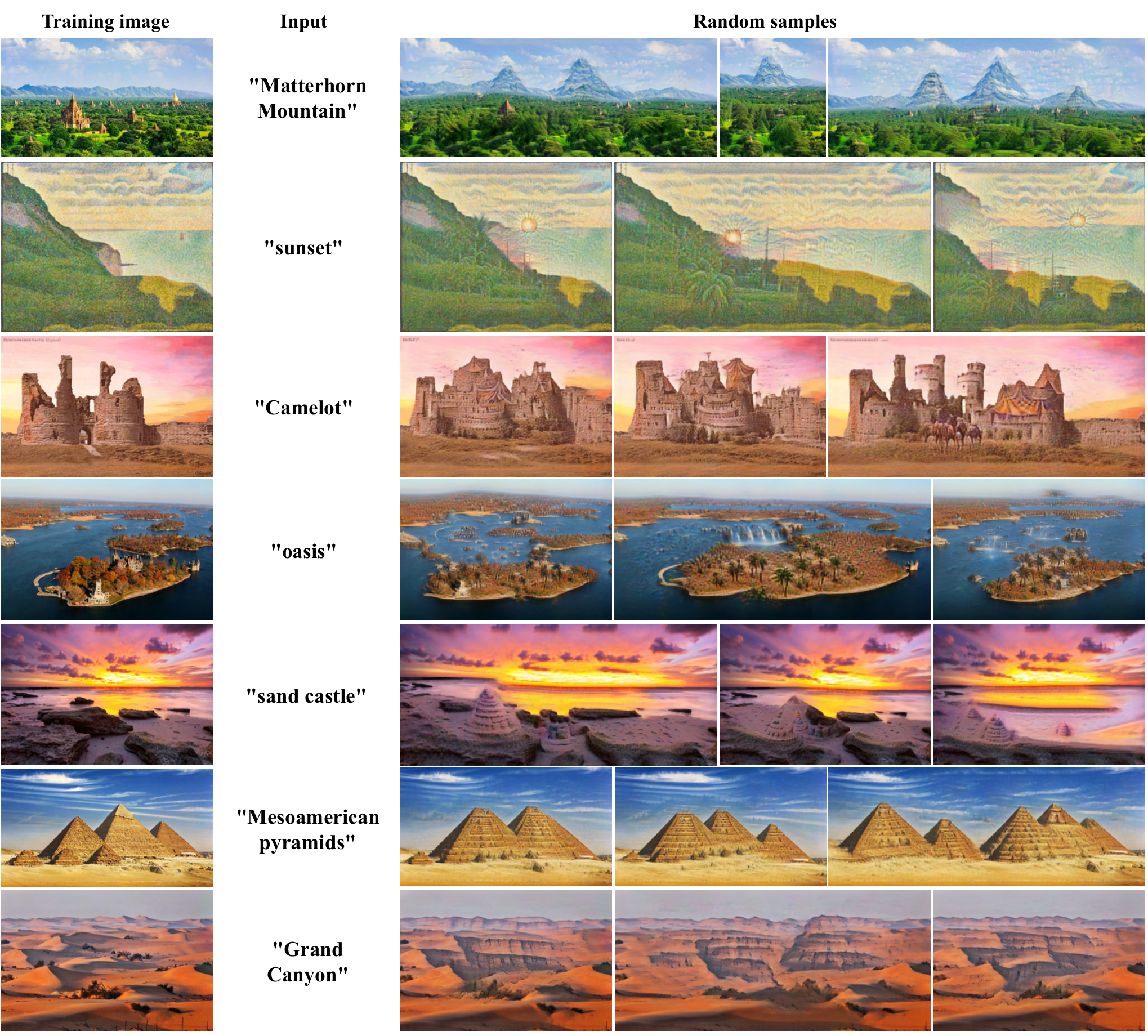}
  \caption{\textbf{Text-guided content generation.}} 
  \label{fig:gen_content_sup}
\end{figure}

\begin{figure}[h]
  \centering
  \includegraphics[width=\textwidth]{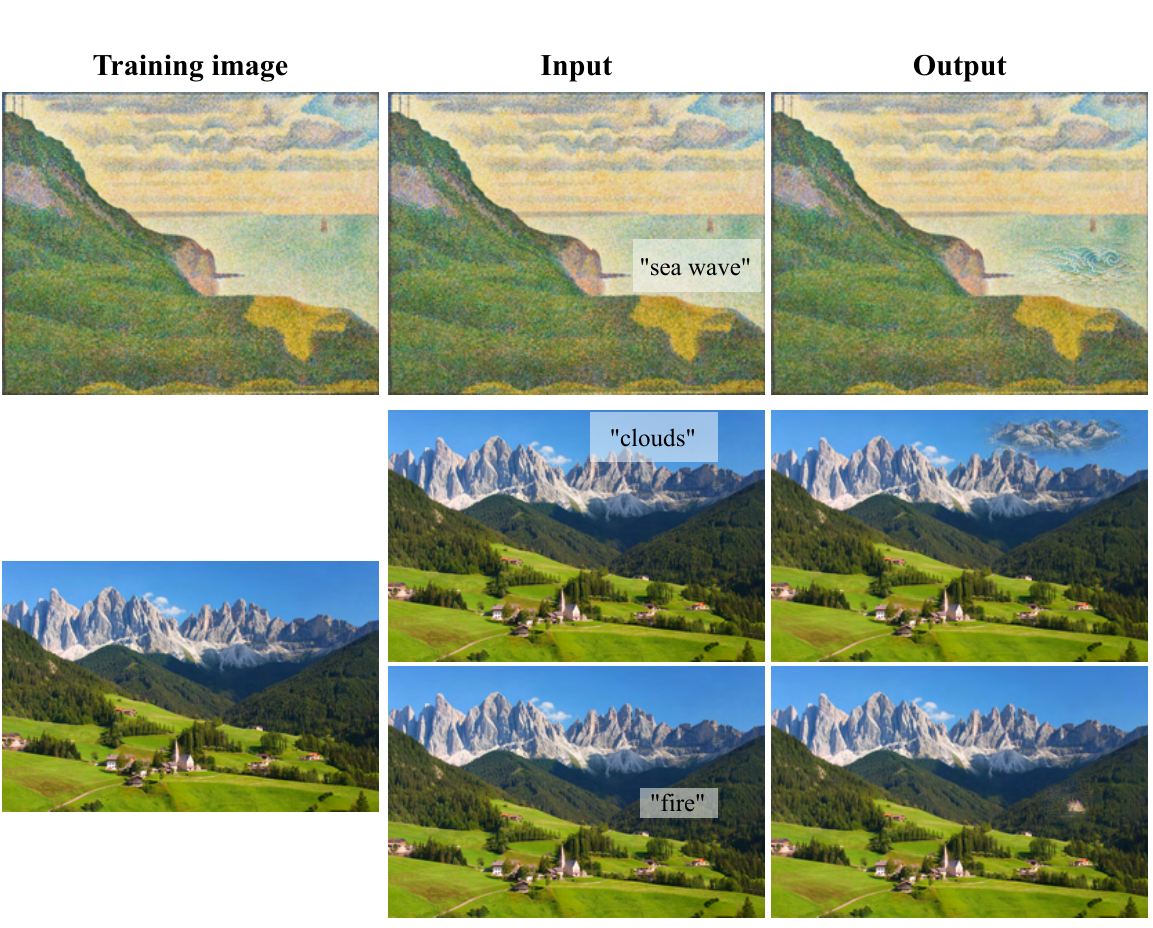}
  \caption{\textbf{Text-guided content generation within a ROI.}}
  \label{fig:clip_roi}
\end{figure}

\begin{figure}[h]
  \centering
  \includegraphics[width=\textwidth]{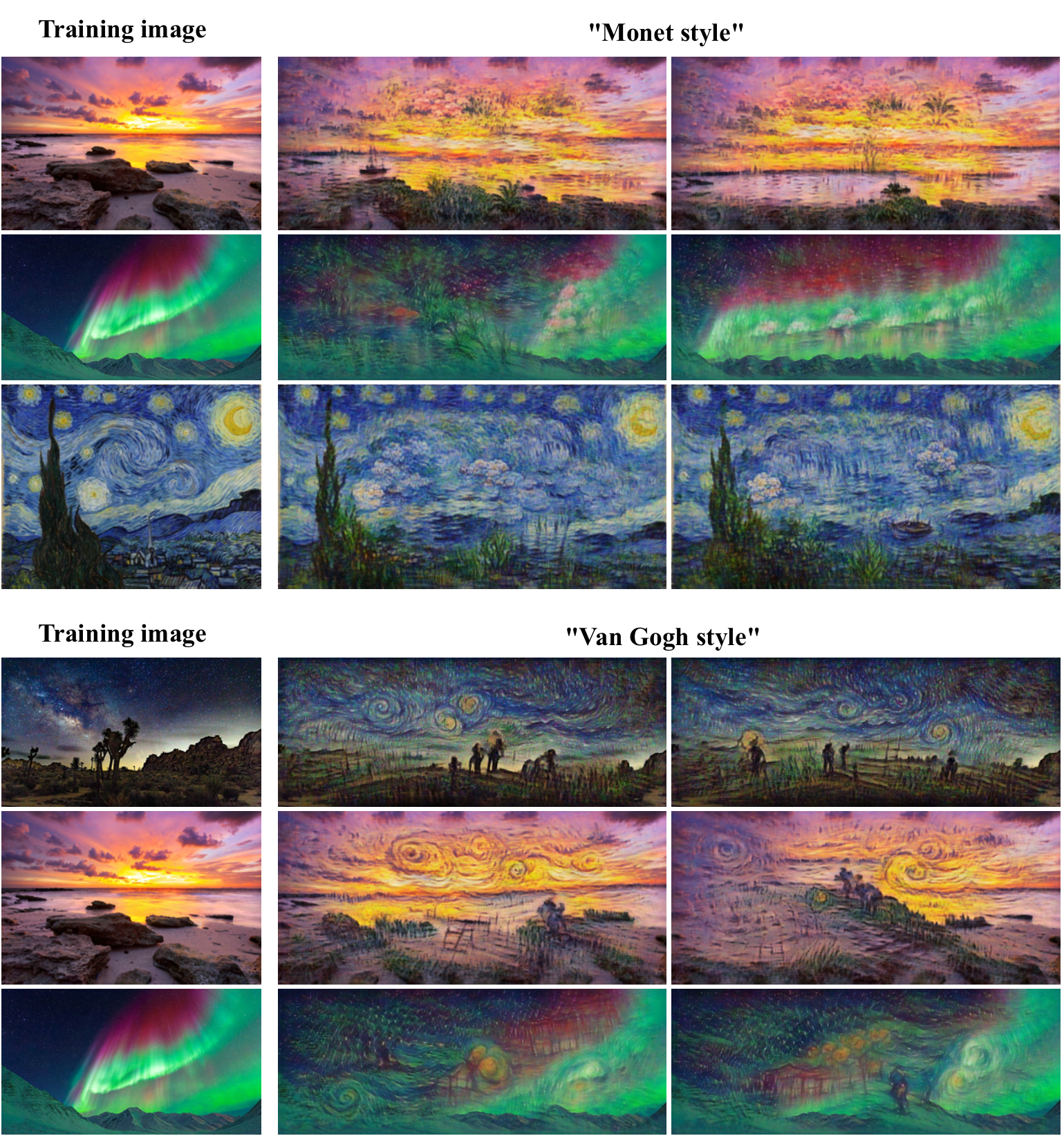}
  \caption{\textbf{Sampling with text-guided style at arbitrary aspect ratios.} 
  }
  \label{fig:style_a_d}
\end{figure}

\begin{figure}[h]
  \centering
  \includegraphics[width=\textwidth]{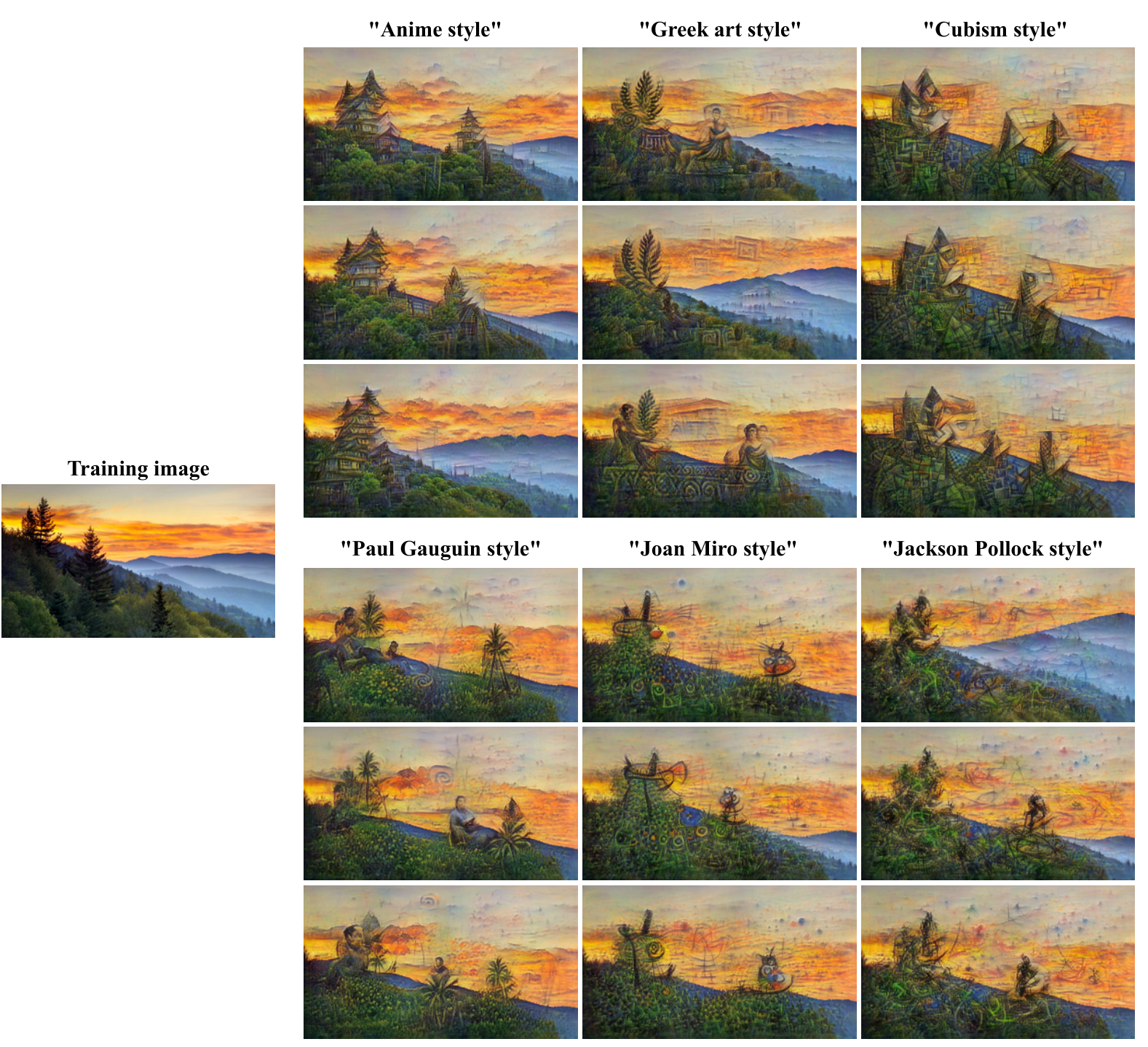}
  \caption{\textbf{Sampling with text guided style.} 
  }
  \label{fig:style2}
\end{figure}

\begin{figure}[h]
  \centering
  \includegraphics[width=\textwidth]{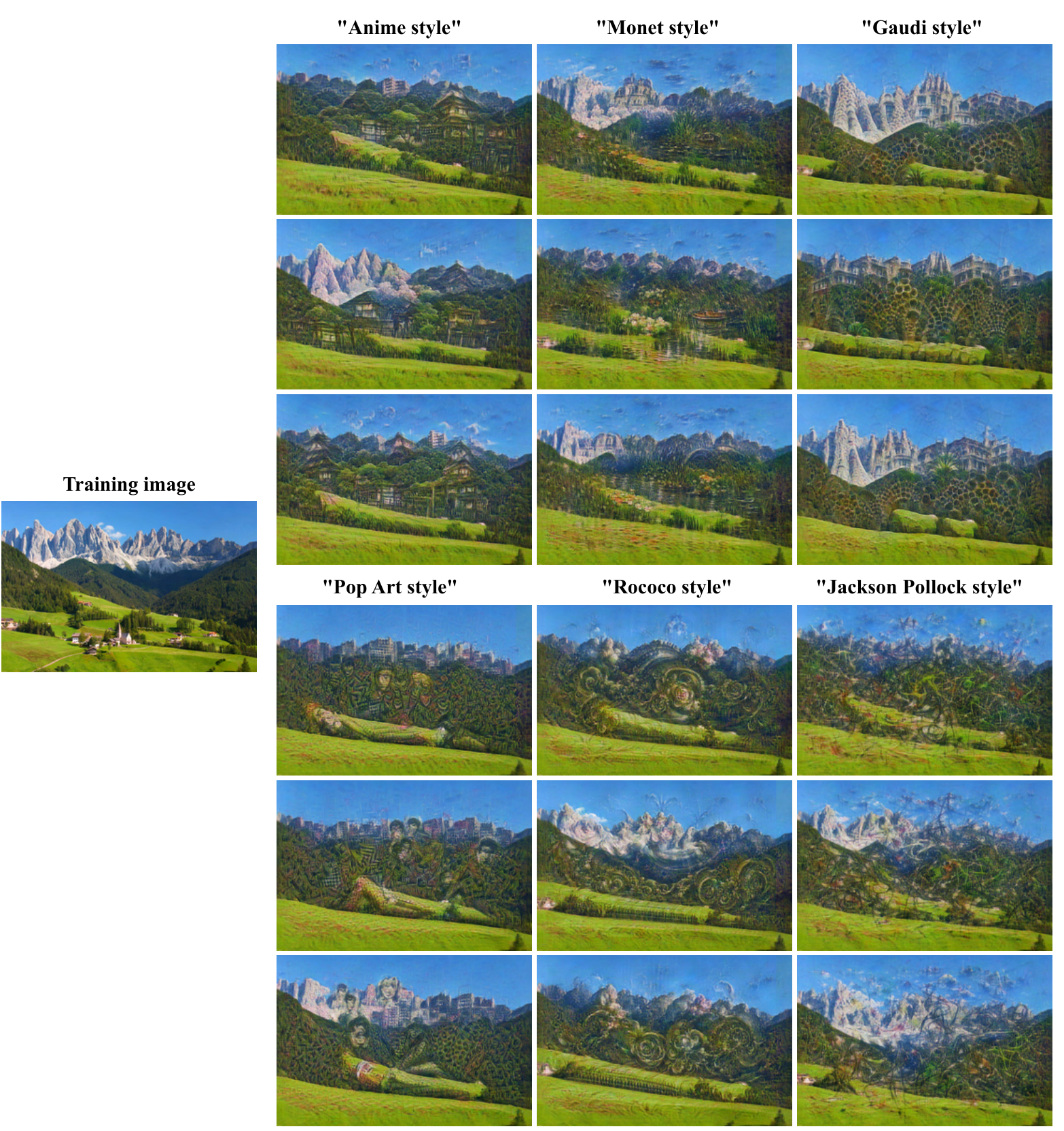}
  \caption{\textbf{Sampling with text guided style.} 
  }
  \label{fig:style3}
\end{figure}

\begin{figure}[h]
  \centering
  \includegraphics[width=\textwidth]{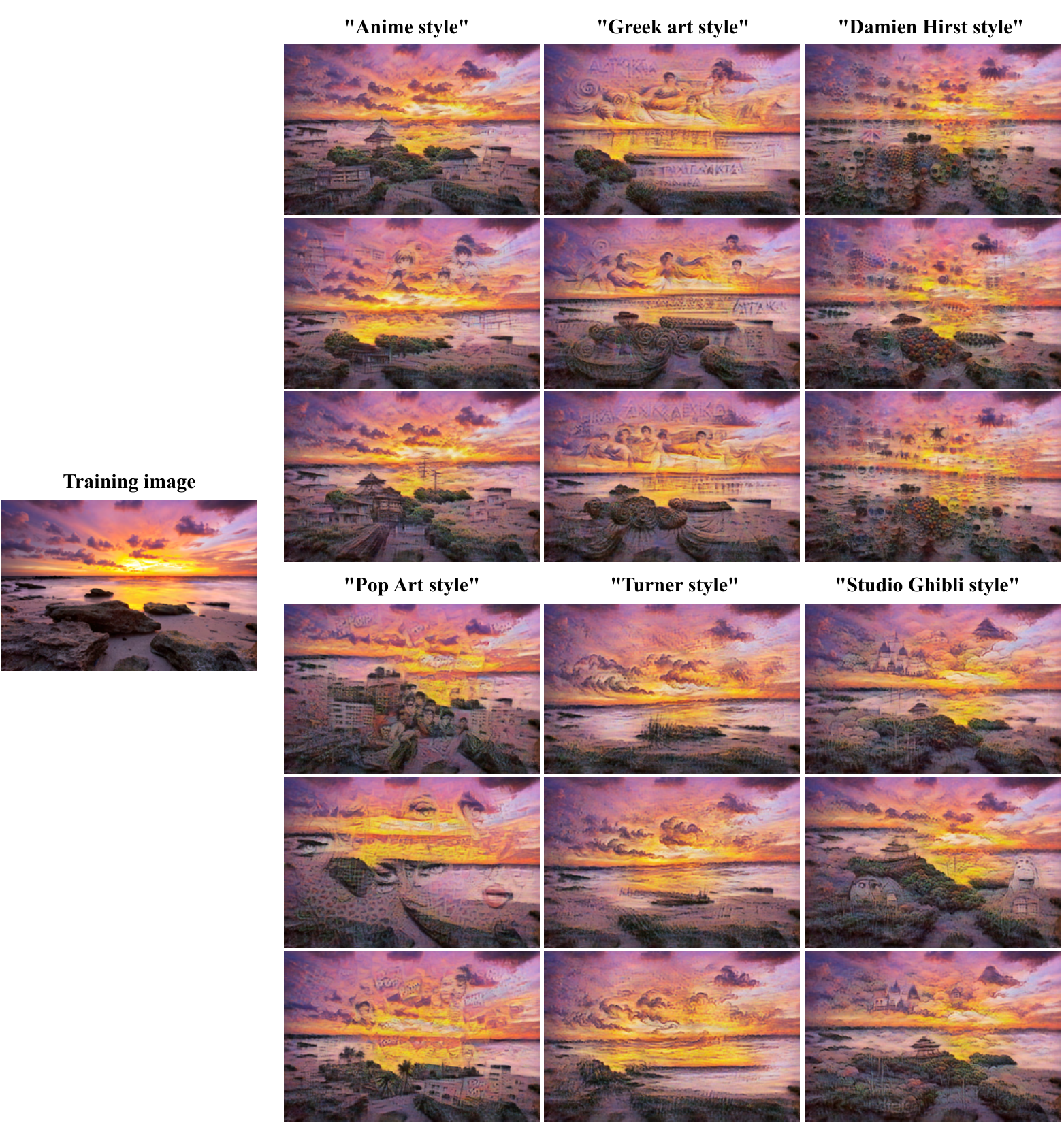}
  \caption{\textbf{Sampling with text guided style.} 
  }
  \label{fig:style4}
\end{figure}

\begin{figure}[h]
  \centering
  \makebox[\textwidth]{
  \includegraphics[width=\textwidth]{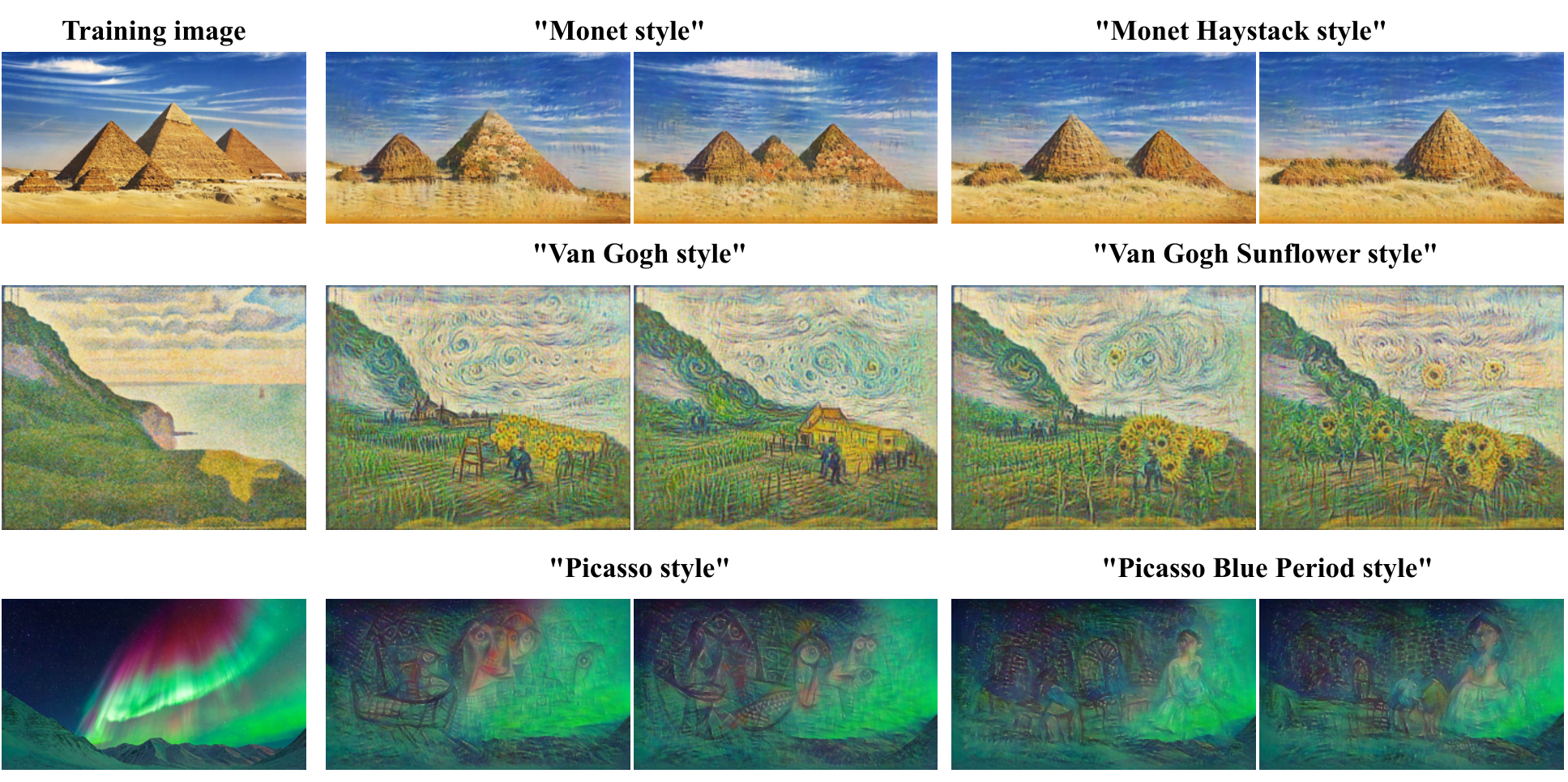}
  }
  \caption{\textbf{Sampling with text guided style using a focused text prompt.} Oftentimes, when specifying a general style, CLIP guides the generation towards the most famous attributes associated with that style. For example, ``Monet style'' usually results in adding flowers. To obtain more focused results, it is possible to add the name of a painting or a distinct period associated with the desired artist. The resulting effect is exemplified on the right. 
  }
  \label{fig:painting}
\end{figure}

\begin{figure}[h]
  \centering
  \includegraphics[width=\textwidth]{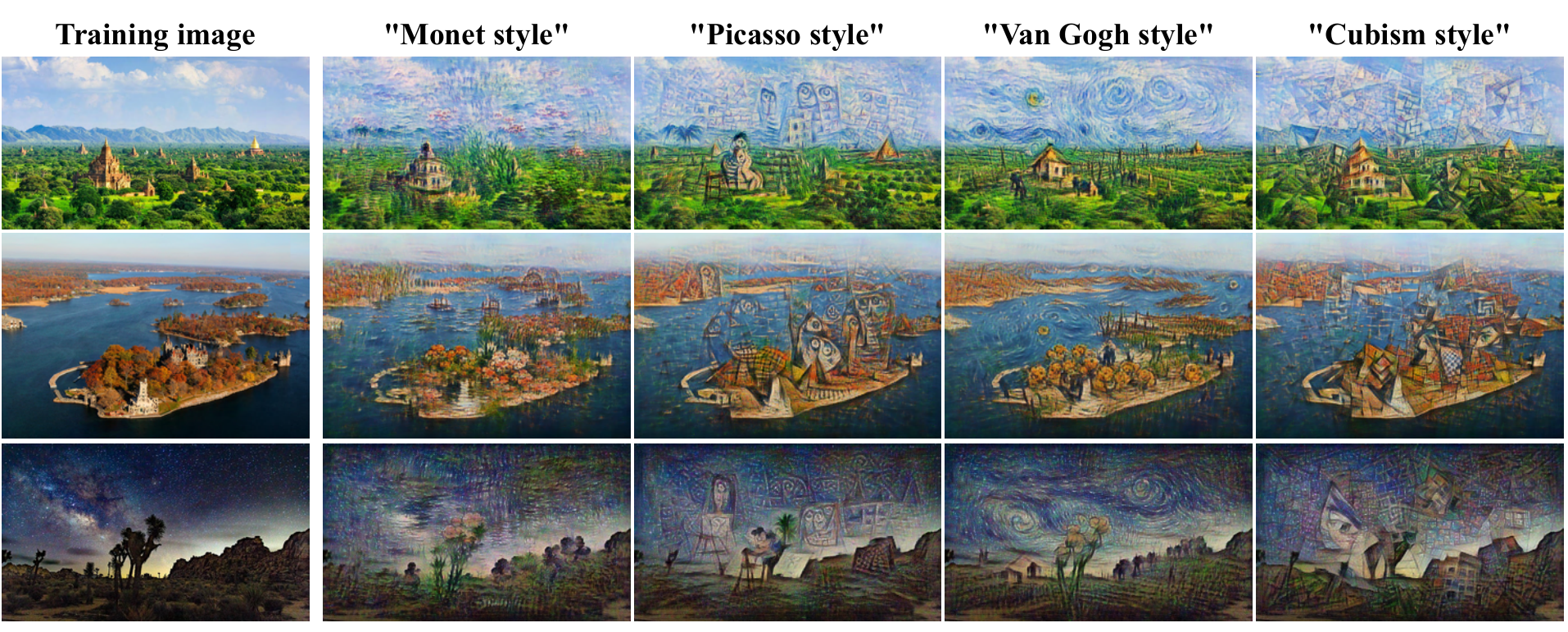}
  \caption{\textbf{Text-guided style transfer.} Rather than controlling the style of the random samples generated by our model, we can also use SinDDM to modify the style of the training image itself. We achieve this by injecting the training image directly to the finest scale 
  so that the modifications imposed by our denoiser and by the CLIP guidance only affect the fine textures. This leads to a style-transfer effect, but where the style is dictated by a text prompt rather than by an example style image.
  }
  \label{fig:starting_from_the_image}
\end{figure}

\begin{figure}[h]
  \centering
  \includegraphics[width=\textwidth]{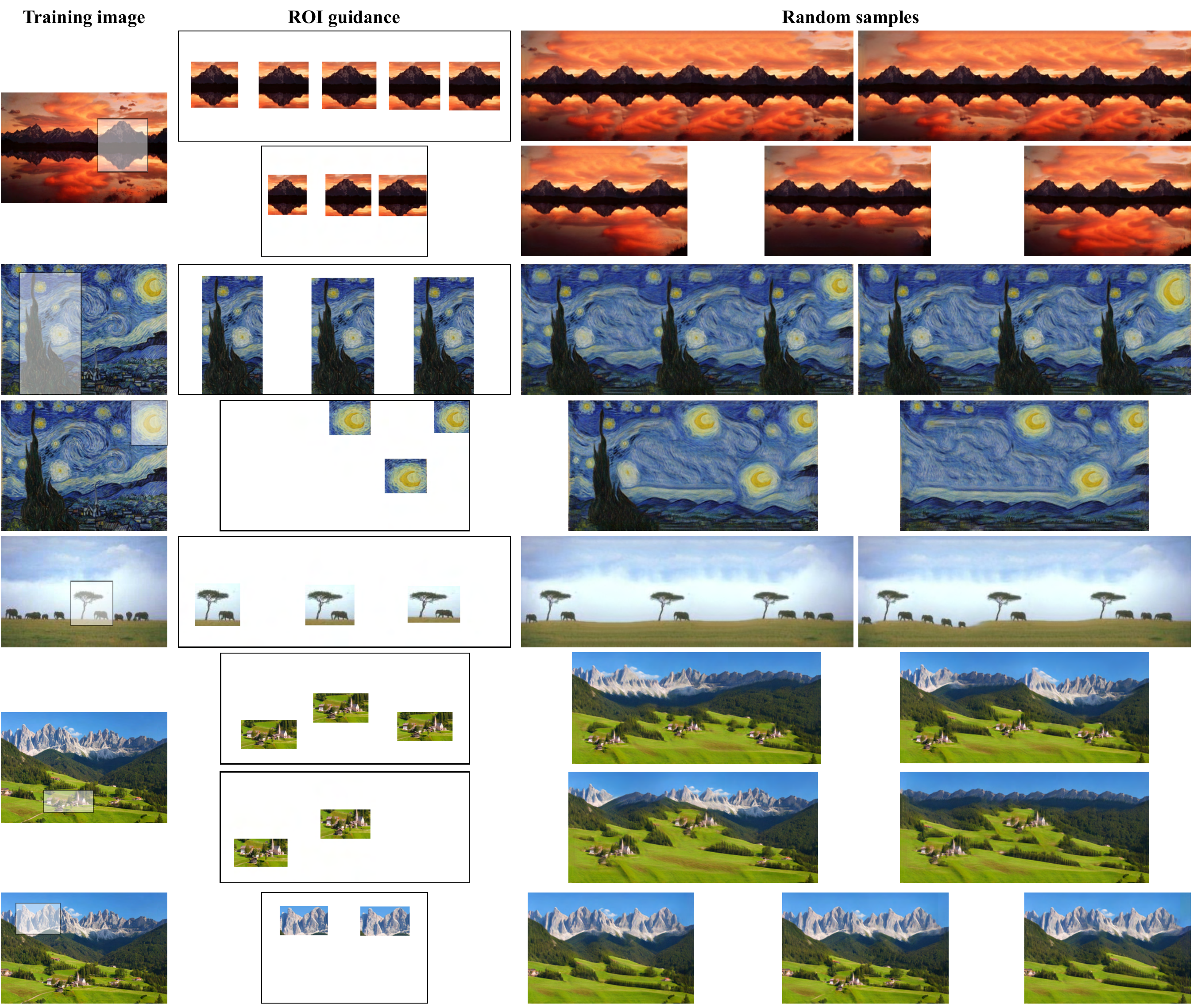}
  \caption{\textbf{Generation guided by image contents within ROIs.}
  }
  \label{fig:ROI_sup}
\end{figure}

We provide additional examples for all the applications described in the main text. \Cref{fig:gen_sup} presents additional examples for unconditional sampling.  In Figs.~\ref{fig:gen_content_sup} and 
\ref{fig:clip_roi} we provide additional examples for image generation with text-guided contents, with and without ROI guidance, respectively. Figures~\ref{fig:style_a_d}, 
\ref{fig:style2}, \ref{fig:style3}, \ref{fig:style4} and \ref{fig:painting} provide additional examples of image generation with text-guided style, and in \cref{fig:starting_from_the_image} we provide examples for text-guided style transfer. Finally, \cref{fig:ROI_sup} illustrates generation guided by image contents within ROIs.

\clearpage

\clearpage

\clearpage

\clearpage
\section{SinDDM Architecture}\label{appendix:Arch}
\begin{figure}[H]
  \centering
  \includegraphics[width=\textwidth]{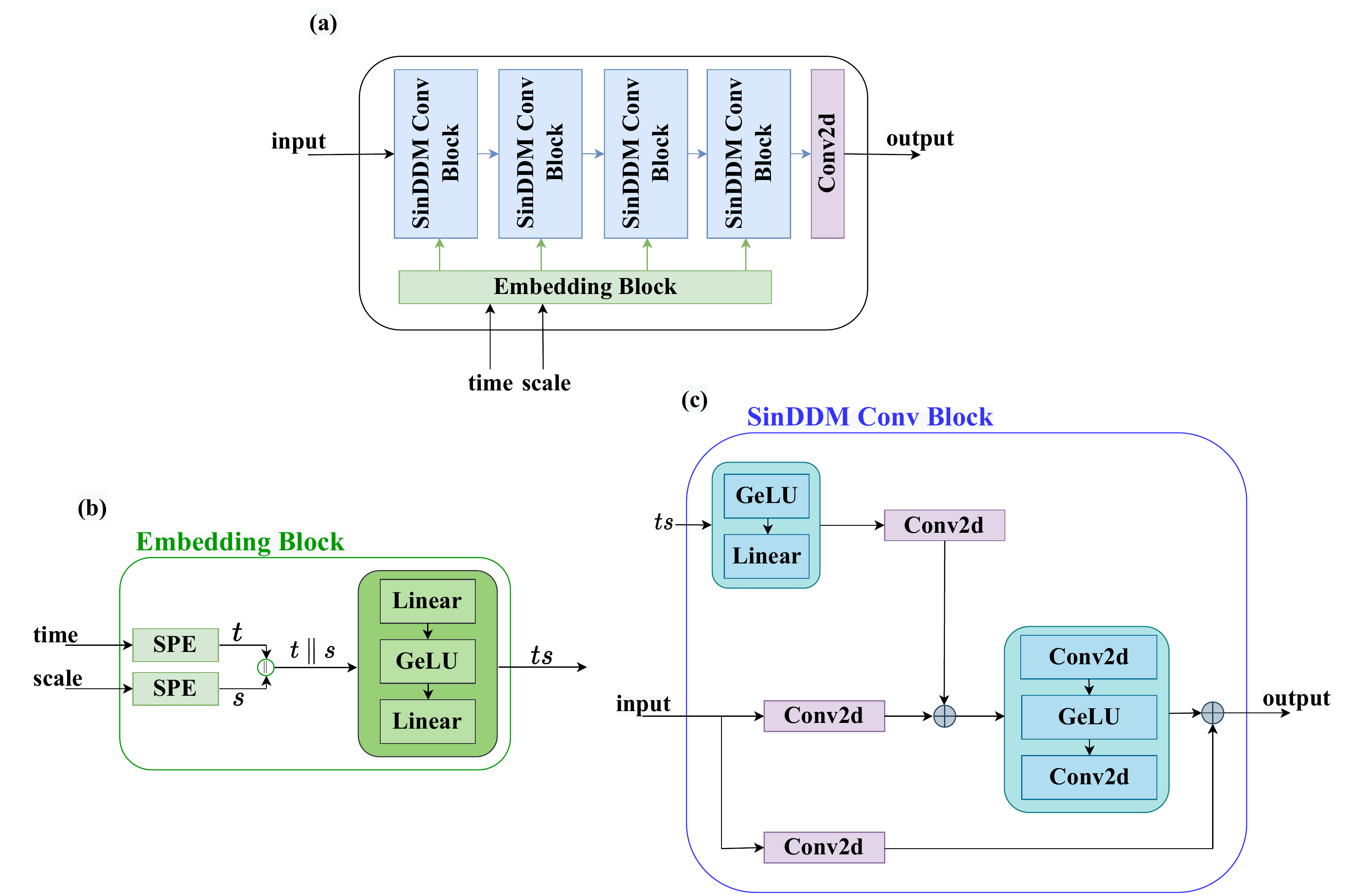}
  \caption{\textbf{SinDDM architecture.} Our model's architecture (a) comprises an embedding block (b) that gets as inputs the timestep $t$ and scale $s$, and a fully convolutional image pipeline (c) that is conditioned on the embedding vector of $t$ and $s$.}
  \label{fig:arch_sup}
\end{figure}

Similarly to externally-trained diffusion models, SinDDM receives a noisy image and a timestep $t$ as inputs, and it predicts the noise. However, unlike external models, in our setting the noisy image is also a bit blurry, and our model also accepts the scale $s$ as input. This is illustrated in \cref{fig:arch_sup}(a). Since we train on a single image, we must take measures to avoid memorization of the image. We do so by limiting the receptive field of the model. 
Specifically, we use a fully convolutional pipeline for the image input, which consists of 4 SinDDM conv blocks (\cref{fig:arch_sup}(c)). Each of these blocks has a receptive field of $9\times 9$, yielding a total receptive field of $35\times 35$. The timestep $t$ and scale $s$ first pass through an embedding block (\cref{fig:arch_sup}(b)), in which they go through Sinusoidal Positional Embedding (SPE) and then concatenated and passed through two fully-connected layers with GeLU activations to yield a time-scale embedding vector $ts$ (\cref{fig:arch_sup}(b)). SinDDM's conv block's data input pipeline consists of 2D convolutions with a residual connection and GeLU activations, and the input time-scale embedding vector is passed through a GeLU activation and a fullly-connected layer. The time-scale embedding vector is passed to each SinDDM conv block as depicted in \cref{fig:arch_sup}(a). 

\section{Training Details}
\label{appendix:Training}
We train our model using the Adam optimizer with its default Torch parameters. Similarly to the original DDPM training scheme, we apply an Exponential Moving Average (EMA) on the model's weights with a decay factor of 0.995.  When using an A6000 GPU, we train our model for $120\times10^3$ steps with an initial learning rate of $0.001$, which is reduced by half on $[20, 40, 70, 80, 90, 110]\times10^3$ steps. We set the batch size to be 32. For a $200\times250$ image, training takes around 7 hours.

We also experimented with training on an RTX2080 GPU, which is slower and has less memory. In this case, we take the batch size to be 16, multiply the number of steps by 4, and modify the scheduler steps accordingly. In this setting, training on a $200\times250$ image takes around 20 hours.

\section{Derivation of the Forward and Reverse Diffusion Processes}\label{appendix:diff_process}
In the sampling process, at each scale except for the coarsest ($s=0$), the reverse diffusion begins with a noisy version of the upsampled image generated at the previous scale. Since the upsampling introduces some blur, this implies that our model's task is to perform both deblurring and denoising. To account for this, our training algorithm for $s>0$ is not identical to the standard one. Specifically, in our case we construct the image $x_t$ at train time as a linear combination of not only the clean training image and Gaussian noise (as in DDPM \citep{ho2020denoising}), but also of an upsampled version of the training image from the previous scale, which is a bit blurry. 

Concretely, at train time we define
\begin{equation}
  x_{t}^{s,\text{mix}} = \gamma^s_{t} \tilde{x}^s + (1 - \gamma^s_{t})  x^{s},
  \label{eq:blurry_mix}
\end{equation}
where $\gamma^s_{t}\in [0,1]$ is a non-increasing monotonic function of $t$ (see below). Thus, $x_{t}^{s,\text{mix}}$ is a linear combination of $x^{s}$ and its blurry version $\tilde{x}^s$. We then construct $x_{t}^{s}$ as 
\begin{equation}
    x^s_t = \sqrt{\bar{\alpha}_t}x^{s,\text{mix}}_t + \sqrt{1-\bar{\alpha}_t}\epsilon,
    \label{reparam_inference_dist}
    \end{equation}
where $\epsilon\sim\mathcal{N}\left(0, I\right)$ and $\bar{\alpha}_{t}$ follows a cosine schedule as in Improved DDPM \citep{nichol2021improved}. Therefore, conditioned on $x^s$ and $\tilde{x}^s$, we have that $x_{t}^{s} \sim \mathcal{N}(\sqrt{\bar{\alpha}_{t}}x^{s,\text{mix}}_t,\sqrt{1-\bar{\alpha}_{t}}I)$. Now, for each timestep $t$, we train our model to predict the noise from $x_{t}^{s}$, as in DDPM \citep{ho2020denoising}. Substituting our noise estimate for $\epsilon$ in (\ref{reparam_inference_dist}), we can isolate $x^{s,\text{mix}}_t$. This is line 6 of the sampling algorithm in the main text. And given $x^{s,\text{mix}}_t$, we can isolate $x^{s}$ from (\ref{eq:blurry_mix}). This leads to the estimate  $\hat{x}_0^s$ appearing in line 7 of the sampling algorithm.


\subsection{The Maximal Noise Level in Each Scale}\label{appendix:noise_level}
We train each scale for $T=100$ timesteps. However, when sampling from the model, we start the generation in every scale $s>0$ from
\begin{equation}
  T[s] = \min \left\{ t \;\;: \;\; \frac{\sqrt{1-\bar{\alpha_{t}}}}{\sqrt{\bar{\alpha_{t}}}} > \left\|x^{s} - \tilde{x}^s\right\|_{2}\right\}.
  \label{eq:sampling}
\end{equation}
This ensures that the effective noise-to-signal ratio that is due to the noise, is roughly the same as the $L^2$ norm between the original image in scale $s$ and its blurry version (the upsampled image from scale $s-1$). In other words, this guarantees that we add just enough noise to generate the missing details in that scale, but not too much to mask the details already generated in the previous scale. 
It should be noted that with this strategy, we train on timesteps that we do not eventually use in sampling (those with $t>T[s]$). However, we found this to lead to better results than training only on the timesteps used in sampling.


\subsection{The $\gamma^s_t$ Schedule}\label{appendix:gamma}
For $s=0$, we set $\gamma_t^s=0$ for all $t$ since we perform only denoising and not deblurring. For every $s>0$ and $t\leq T[s]$ we choose $\gamma^s_{t}$ such that the noise level is similar to the blur level. Thus, in the reverse diffusion, each step performs denoising just as it does deblurring. 
We do this by choosing $\gamma^s_t$ such that
\begin{equation}
  \left\|x^{s} - x_{t}^{s,\text{mix}}\right\|_{2} = \frac{\sqrt{1-\bar{\alpha_{t}}}}{\sqrt{\bar{\alpha_{t}}}}. 
  \label{eq:gamma_t}
\end{equation}
This equation possesses a closed form solution (obtained by substituting (\ref{eq:blurry_mix}) in (\ref{eq:gamma_t})), given by
\begin{equation}
\gamma^s_{t}=\frac{1}{\left\|x^{s} - \tilde{x}^s\right\|_{2}} \frac{\sqrt{1-\bar{\alpha_{t}}}}{\sqrt{\bar{\alpha_{t}}}}.
\end{equation}
Recall that during training we also work with $t>T[s]$, in which case the equation would give $\gamma^s_t>1$. Therefore, at train time we use $\gamma^s_t=1$ for $t>T[s]$. During sampling, we clamp $\gamma^s_t$
 to $0.55$ for all timesteps and scales.

To illustrate the necessity of adding blur during training we trained models using the aforementioned $\gamma^s_t$ schedule with $\gamma^s_t=0,\forall s,t$, meaning without blur. Removing the blurring from the training results in smoother samples with lack of fine details. A qualitative comparison between models trained with and without blur is shown in \cref{fig:blur_comp}.

\subsection{The Sampling Process}\label{appendix:sampling_process}
As shown in the DDIM formulation \citep{song2020denoising}, in a regular diffusion process, the distribution $q(x_{t-1}|x_{t}, x_{0})$ can be chosen as
\begin{equation}
    q_{\sigma}(x_{t-1}|x_{t},x_{0}) = \mathcal{N}\left(\sqrt{\bar{\alpha}_{t-1}}x_0 + \sqrt{1-\bar{\alpha}_{t-1}-\sigma^2_t}\cdot \frac{x_t-\sqrt{\bar{\alpha}_t}x_0}{\sqrt{1-\bar{\alpha}_{t}}}, \sigma^{2}_{t}\mathbf{I}\right),
    \label{eq:general_infrence_dist}
\end{equation}
where $\sigma_t$ is a hyperparemter. 
In our case, this translates to
\begin{equation}
    q_\sigma(x^s_{t-1}|x^s_{t}, x^s_0) = \mathcal{N}\left(\sqrt{\bar{\alpha}_{t-1}}x^{s,\text{mix}}_{t-1} + \sqrt{1-\bar{\alpha}_{t-1}-\sigma^2_t}\cdot \frac{x^s_t-\sqrt{\bar{\alpha}_t}x^{s,\text{mix}}_{t}}{\sqrt{1-\bar{\alpha}_{t}}}, \sigma^2_t\mathbf{I}\right).
    \label{eq:our_general_infrence_dist}
\end{equation}
We found empirically that the best results are achieved with $\sigma_t^s=\sqrt{\frac{1-\bar{\alpha}_{t-1}}{1-\bar{\alpha}_{t}}}\cdot\sqrt{1-\alpha_t}$ for $s=0$ and $\sigma_t^s=0$ for $s>0$. As described in \citep{song2020denoising}, the first choice corresponds to the DDPM sampling process \citep{ho2020denoising}, while the second choice corresponds to a deterministic process.


\section{Text Guided Generation}\label{appendix:text_gen}
\subsection{Algorithm}\label{appendix:algo}
As described in the main text, text guidance is achieved by adding CLIP's gradients to $\hat{x}^s_0$ in each step, using an adaptive step size. Specifically, we aim at updating $\hat{x}^s_0$ as
\begin{equation}
    \hat{x}_{0}^{s} \leftarrow \eta\; \delta \; m^s \odot \nabla_{\hat{x}^s_0}\mathcal{L}_{\text{CLIP}}  + 
  (1-m^s)\odot \hat{x}_{0}^{s},
  \label{eq:adaptive_step_size}
\end{equation} 
where $\delta=\|\hat{x}_{0}^{s} \odot m\|/\| \nabla_{\hat{x}^s_0}\mathcal{L}_{\text{CLIP}} \odot m \|$ and $\eta \in [0,1]$ is a \emph{strength} parameter that controls the intensity of the CLIP guidance. 
However, due to the strong prior of our denoiser (which is overfit to the statistics of the training image) such update steps tend to be ineffective. This is because each denoiser step undoes the preceding CLIP step. To overcome this effect, we use a momentum over $\hat{x}^s_0$. Specifically, our update rule for $\hat{x}^s_0$ is
\begin{equation}
    \hat{x}_{0}^{s} \leftarrow \eta\; \delta \; m^s \odot \nabla_{\hat{x}^s_0}\mathcal{L}_{\text{CLIP}}  + 
  (1-m^s)\odot \left(\lambda\hat{x}_{0}^{s} + (1-\lambda)\hat{x}_{0,\text{prev}}^{s}\right),
  \label{eq:update_rule}
\end{equation} 
where $\hat{x}_{0,\text{prev}}^{s}$ is $\hat{x}_{0}^{s}$ from the previous timestep and $\lambda\in[0,1]$ is a momentum parameter which we set to $0.05$ in all our experiments.
Additionally, in contrast to regular sampling, here we use $\gamma_t^s=0$ for all $t$ and $s$ and $\sigma_t=\sqrt{\frac{1-\bar{\alpha}_{t-1}}{1-\bar{\alpha}_{t}}}\cdot\sqrt{1-\alpha_t}$ for every $s$. In other words, we use the DDPM sampling algorithm \citep{ho2020denoising} for all scales, which only denoises the image and does not explicitly attempt to remove blur.

\subsection{Data Augmentation for Text Guided Generation}\label{appendix:data_aug}
Following \citep{bar2022text2live}, we augment our images and text, feeding all augmented inputs into CLIP's image encoder and text encoder. 
For regular text-guidance, we use the image and text augmentations from \citep{bar2022text2live}. For text-guidance within a ROI, we feed only the ROI (and its augmentations) into CLIP's image encoder. This requires upsampling the ROI to the size of $224\times 224$, which typically results in a very blurry image. Therefore, in this case, we additionally augment the given text prompt using the following text templates:
\begin{itemize}
\item ``photo of \{\}.'' 
\item  ``low quality photo of \{\}.'' 
\item  ``low resolution photo of \{\}.''
\item  ``low-res photo of \{\}.''
\item  ``blurry photo of \{\}.''
\item  ``pixelated photo of \{\}.''
\item  ``a photo of \{\}.''
\item  ``the photo of \{\}.''
\item  ``image of \{\}.''
\item  ``an image of \{\}.''
\item  ``low quality image of \{\}.''
\item  ``a low quality image of \{\}.''
\item  ``low resolution image of \{\}.''
\item  ``a low resolution image of \{\}.''
\item  ``low-res image of \{\}.''
\item  ``a low-res image of \{\}.''
\item  ``blurry image of \{\}.''
\item  ``a blurry image of \{\}.''
\item  ``pixelated image of \{\}.''
\item  ``a pixelated image of \{\}.''
\item  ``the \{\}.''
\item  ``a \{\}.''
\item  ``\{\}.''
\item  ``\{\}''
\item  ``\{\}!''
\item  ``\{\}\ldots''
\end{itemize}

\subsection{Effect of Initial Scale in Generation with Text Guided Contents}\label{appendix:initial_scale}
\begin{figure}[H]
  \centering
  \includegraphics[width=\textwidth]{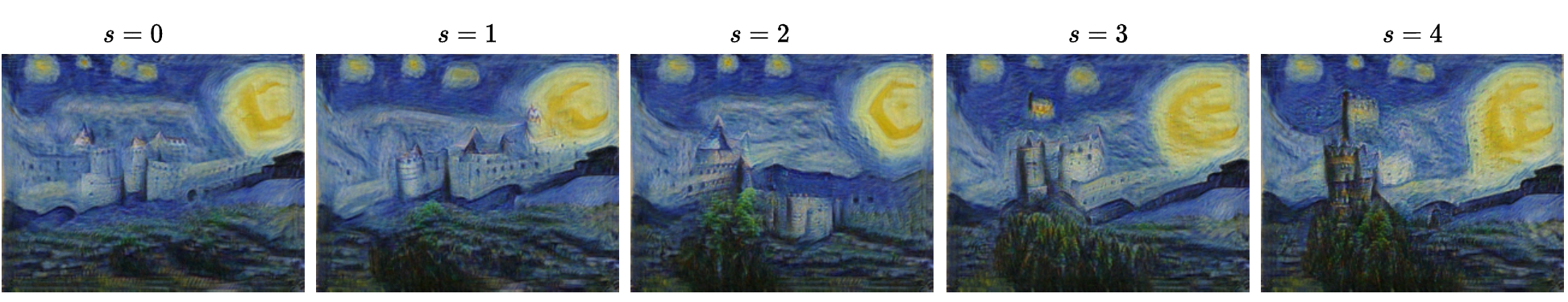}
  \caption{\textbf{Effect of initial scale on generation with text guided content.} Here, the training image is Van Gogh's ``Starry Night'', the text prompt is ``medieval castle'', the fill-factor is $f=0.3$ and the strength is $\eta=0.3$. When applying CLIP guidance only in finer scales, the global structure of the image is already set (by the previous scales) and the guidance only changes fine details and textures. When starting from coarse scales, CLIP's guidance modifies also the global structure.
  }
  \label{fig:clip_scale}
\end{figure}

In text-guided image generation, we can choose the scale from which to begin the guidance. All the results in the main text are with CLIP guidance from the second-coarsest scale, $s=1$. The effect of the initial scale is illustrated in \cref{fig:clip_scale}. When using CLIP guidance from $s=0$, we take the first $T/2$ steps to be purely generative, without any guidance. This allows the model to create diverse global structures before incorporating the CLIP guidance. 

\clearpage
\subsection{Effect of the Fill-Factor and Strength Parameters}\label{appendix:fill_factor}
\begin{figure}[H]
  \centering
  \includegraphics[width=\textwidth]{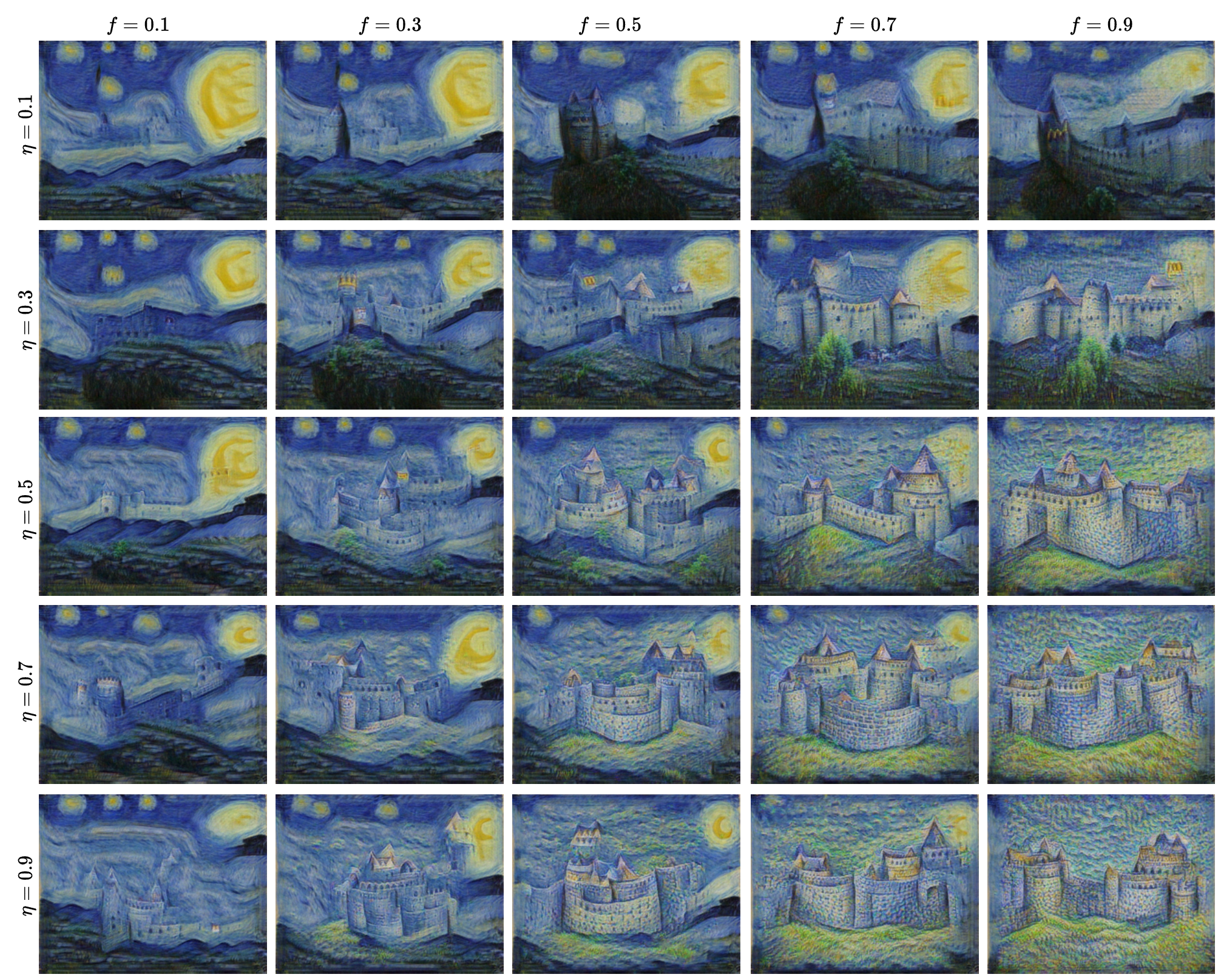}
  \caption{\textbf{Effect of user prescribed parameters on generation with text guided content.} Here, the training image is Van Gogh's ``Starry Night'', the text prompt is ``medieval castle'', and the starting scale is $s=1$. As the fill factor $f$ increases, a larger region of the image is affected by the text. As the strength $\eta$ increases, the modification within the affected regions is more prominent.}
  \label{fig:clip_params}
\end{figure}

We let the user choose the values of the fill factor $f$ and the strength parameter $\eta$, in order to achieve the desired result. The effects of these parameters are illustrated in \cref{fig:clip_params}.  

\section{Controlling Object Sizes}\label{appendix:ObjectSize}
\begin{figure}[H]
  \noindent
  \includegraphics[width=\textwidth]{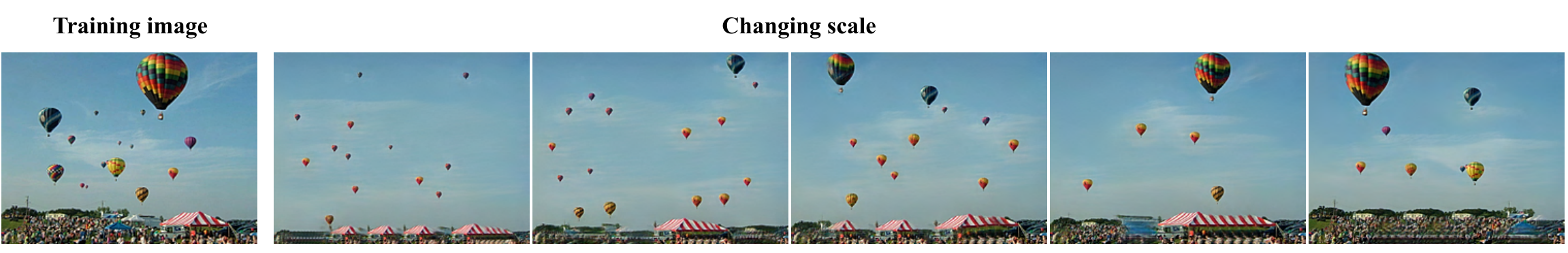}
  \caption{\textbf{Controlling object sizes.} We show the effect of the image dimensions we use relative to the conditioning $s$ with which we supply our model. From left to right, we start from the size of scale 4,3,2,1,0.}
  \label{fig:size}
\end{figure}

Manipulation of image content can be achieved by choosing the dimensions of the initial noise map at the coarsest scale, $s=0$. For example, we can start from a noise map that has the same dimensions as the image at $s=1$. After completing the reverse diffusion process for $s=0$, we pass the output to $s=1$ \emph{without upsampling}, as the image is already at the correct size for that scale. This causes the objects in the image to appear smaller. This is because when injecting to our model the conditioning $s=0$, it generates objects from the smallest training image. Thus, the larger the starting noise map, the smaller the objects appear relative to the image size. This is illustrated in \cref{fig:size}.

\section{Comparisons}\label{appendix:Comparison}
\begin{figure}[H]
  \centering
  \includegraphics[width=0.75\linewidth]{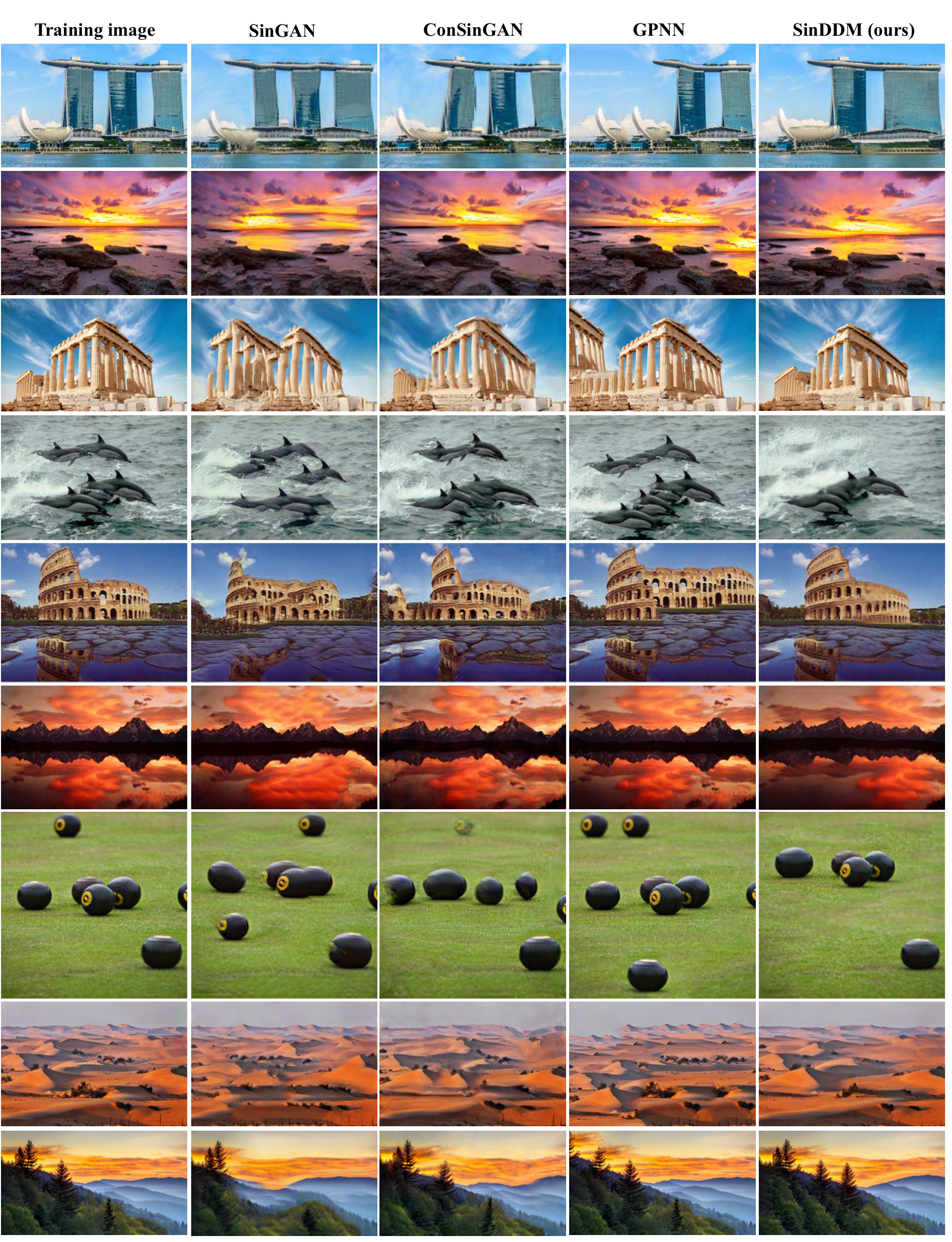}
  \caption{\textbf{Comparison between single-image generative models on the task of unconditional sampling.} We show random samples generated by SinGAN, ConSinGAN, GPNN and SinDDM (ours). Our results are at least on par with existing models in terms of visual quality and generalization beyond the details and object sizes appearing in the training image.}
  \label{fig:gen_comp_sup}
\end{figure}

We next provide comparisons to several competing methods on the tasks of unconditional sampling learned from a single image (\cref{fig:gen_comp_sup}), image generation with text-guided contents (Figs.~\ref{fig:clip_comp_sup1}, \ref{fig:clip_comp_sup2} and \ref{fig:clip_comp_sup3}), image generation with text-guided contents within ROIs (\cref{fig:clip_roi_comp}), image generation with text-guided style (Figs.~\ref{fig:style_comp}, \ref{fig:style_comp1} and \ref{fig:style_comp2}), style transfer (\cref{fig:style}) and harmonization (\cref{fig:harmo}).
\clearpage
\subsection{Unconditional Sampling}\label{appendix:comp_sampling}

\clearpage
\subsection{Generation with Text-Guided Content}\label{appendix:comp_content}
\begin{figure}[H]
  \noindent
\includegraphics[width=\textwidth]{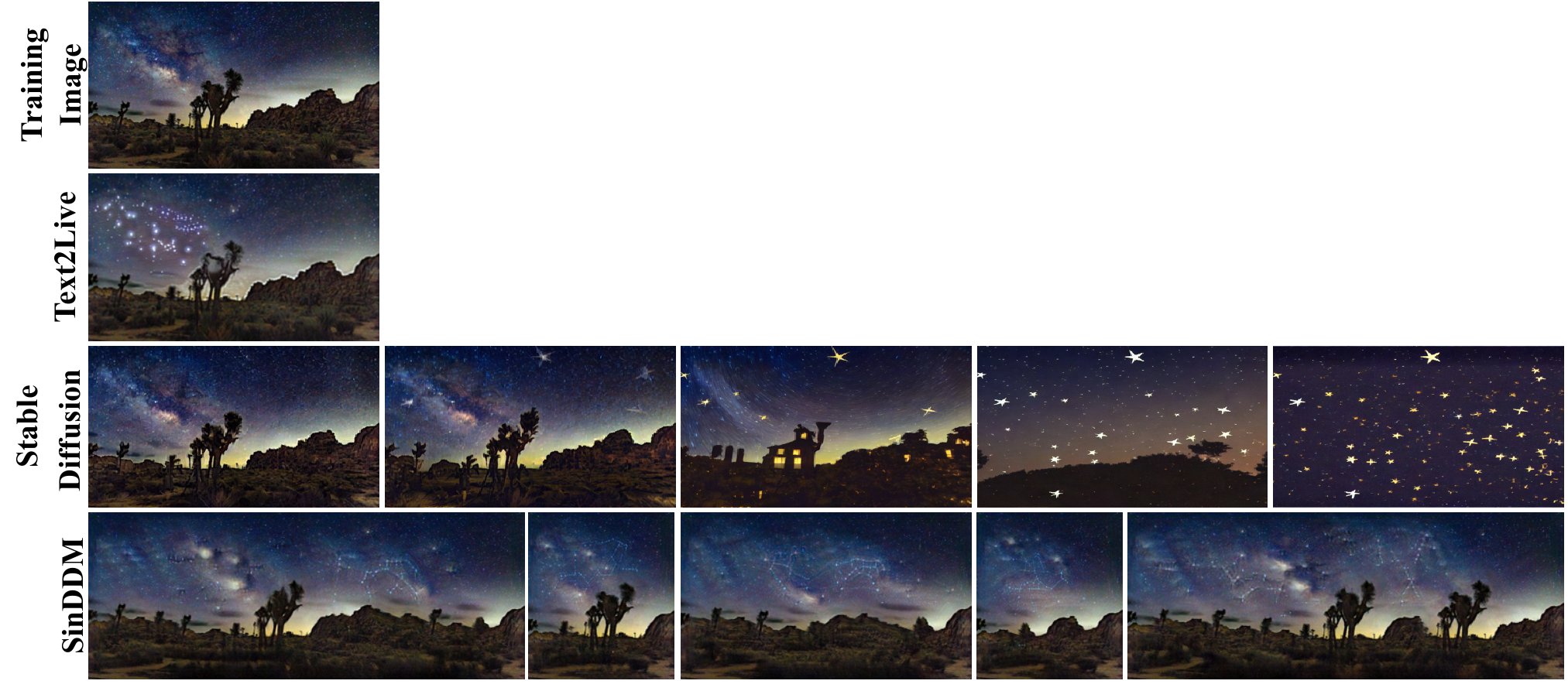}
  \caption{\textbf{Comparisons for image generation and editing guided by text.} Here we used text prompt ``stars constellations in the night sky''. For Text2Live we provided ``stars constellations in the night sky'' as the text describing the edit layer, ``a desert in night with stars constellations in the night sky'' as the text describing the full edited image, 
``a desert in night'' as the text describing the input image, and ``night sky'' as the text describing the region of interest in the input image. For Stable Diffusion we show strengths from 0.2 (left) to 1 (right) in jumps of 0.2.}
  \label{fig:clip_comp_sup1}
\end{figure}

\begin{figure}[H]
  \noindent
\includegraphics[width=\textwidth]{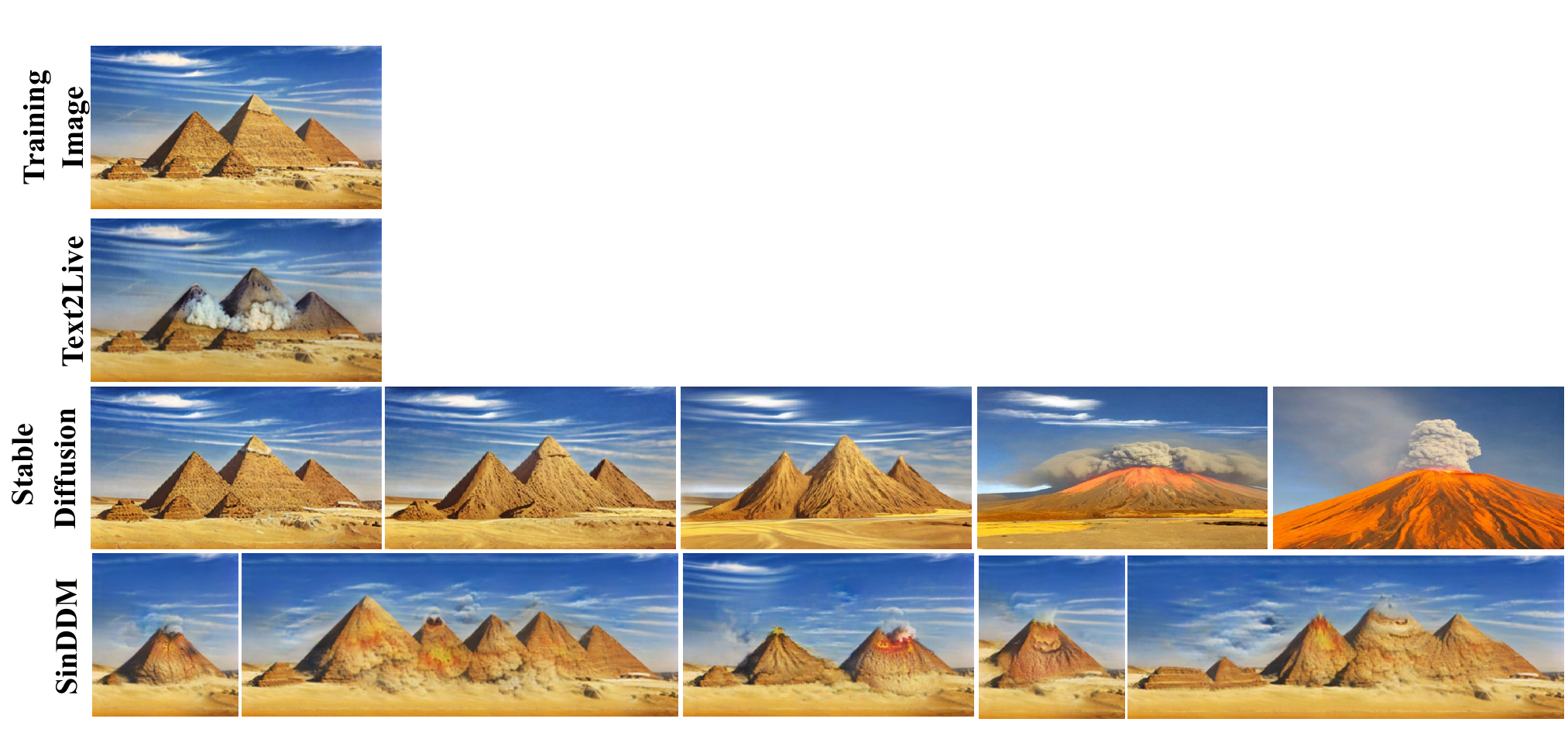}
  \caption{\textbf{Comparisons for image generation and editing guided by text.} Here we used the text prompt ``volcano eruption''. For Text2Live we provided ``volcano eruption'' as the text describing the edit layer, ``volcano erupt from the pyramids in the desert'' as the text describing the full edited image, ``pyramids in the desert'' as the text describing the input image, and ``the pyramids'' as the text describing the region of interest in the input image. For Stable Diffusion we show strengths from 0.2 (left) to 1 (right) in jumps of 0.2.}
  \label{fig:clip_comp_sup2}
\end{figure}

\begin{figure}[H]
  \noindent
\includegraphics[width=\textwidth]{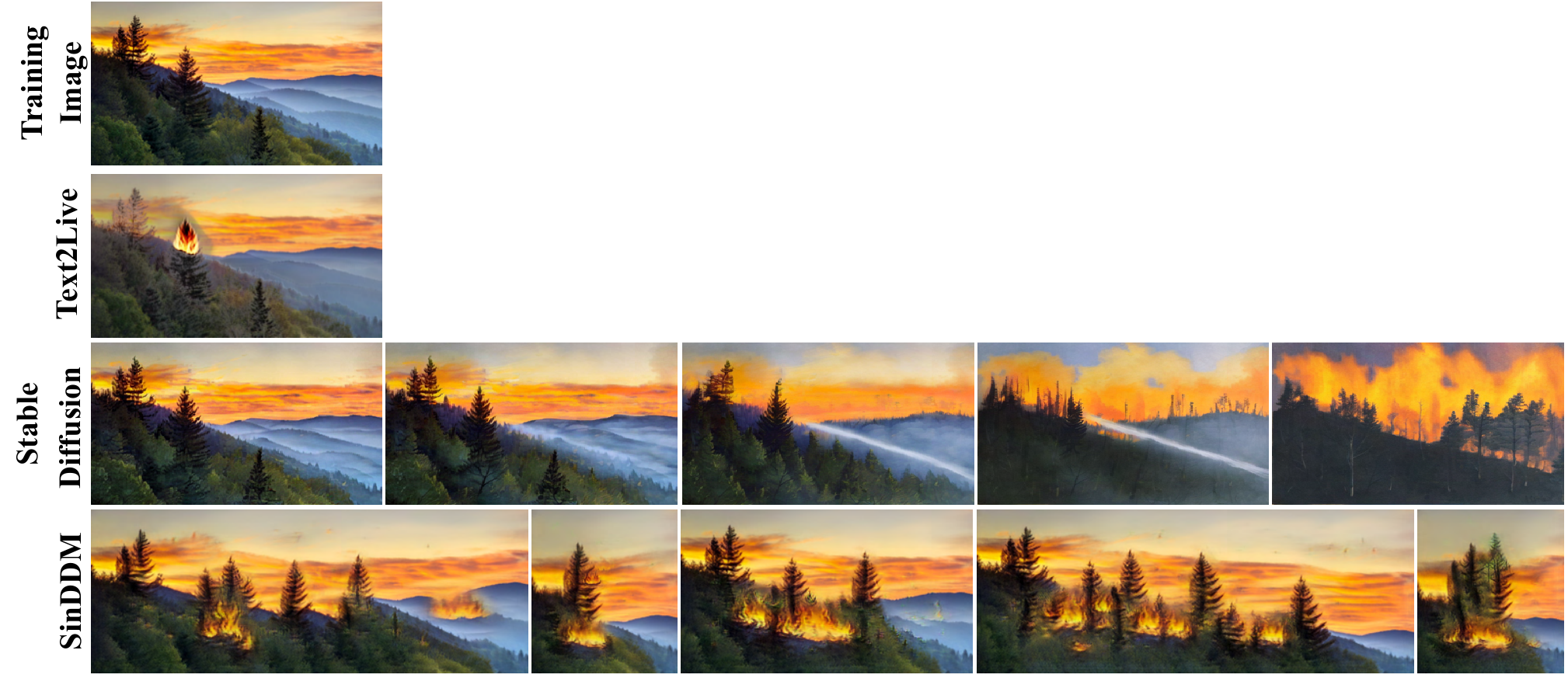}
  \caption{\textbf{Comparisons for image generation and editing guided by text.} Here we used the text prompt ``a fire in the forest''. For Text2Live we provided ``a fire in the forest'' as the text describing the edit layer, 
``a fire in the forest on a mountain in sunset'' as the text describing the full edited image, ``a forest on a mountain in sunset'' as the text describing the input image, and ``the forest'' as the text describing the region of interest in the input image. For Stable Diffusion we show strengths from 0.2 (left) to 1 (right) in jumps of 0.2.}
  \label{fig:clip_comp_sup3}
\end{figure}

We next compare our method to Text2Live \citep{bar2022text2live} and Stable Diffusion \citep{rombach2022high} on the task of text-guided image generation. The comparisons are shown in Figs.~\ref{fig:clip_comp_sup1}, \ref{fig:clip_comp_sup2} and \ref{fig:clip_comp_sup3}. As opposed to existing methods, SinDDM is not constrained to the aspect ratio or precise object configurations of the input image. Therefore, for our method we show several samples at different aspect ratios.

\paragraph{Text2Live} For Text2Live, we used 1000 bootstrap epochs. Furthermore, this method requires four text prompts (three prompts in addition to the one  describing the final desired result, as in our method). In each example, we list all four prompts.

\begin{figure}[H]
  \noindent
\includegraphics[width=\textwidth]{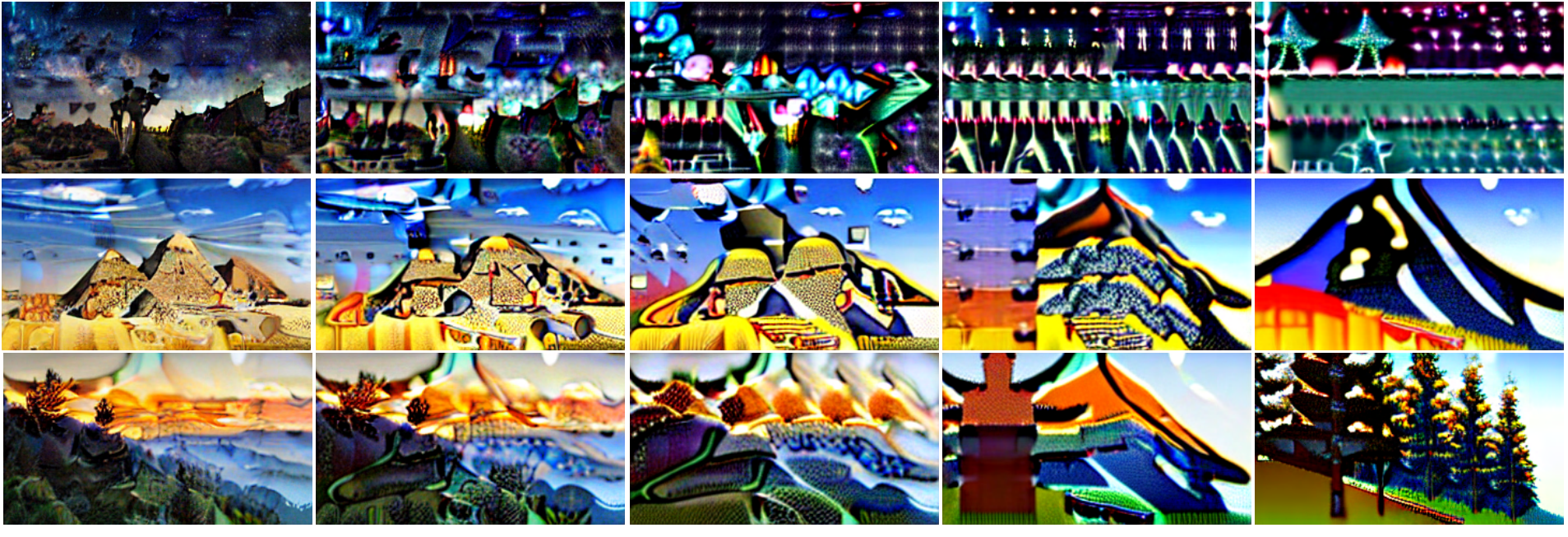}
  \caption{\textbf{Stable diffusion official implementation results.} When using our images in the image-to-image option of the official code of Stable Diffusion, we get poor non-photo-realistic results. To obtain the good results we present in Figs.~\ref{fig:clip_comp_sup1}, \ref{fig:clip_comp_sup2} and \ref{fig:clip_comp_sup3}, we had to use a different resizing strategy before injecting the images into the Stable Diffusion pipeline (see text for details). Here we show the results with official code (without our modification) with strengths from 0.2 (left) to 1 (right) in jumps of 0.2.}
  \label{fig:stable_official}
\end{figure}

\paragraph{Stable Diffusion} For Stable Diffusion, we used the same text we provided to our method. We ran the image-to-image option with different strength values. In each example we show strengths from $0.2$ (left) to~$1$ (right) in jumps of $0.2$. When the strength parameter is small, the edited image is loyal to the original image but the effect of the text is barely noticeable. When the strength parameter is large, the effect of the text is substantial, but the result is no longer similar to the original image. It should be noted that the official code of Stable Diffusion\footnote{\url{https://github.com/CompVis/stable-diffusion}} yields unnatural results for the images we used for comparions (Fig. \ref{fig:stable_official}). We identified that this happens because of the logic which the code uses to scale the image prior to injecting it to the encoder. To overcome this issue, we instead resized the images to $512\times512$ before injecting them to the Stable Diffusion pipeline, and at the output of the pipeline, we scaled the results back to the original dimensions. This leads to natural looking results.

\clearpage
\subsection{Generation with Text-Guided Content in ROI}\label{appendix:comp_roi}
\begin{figure}[H]
  \noindent
  \includegraphics[width=0.9\textwidth]{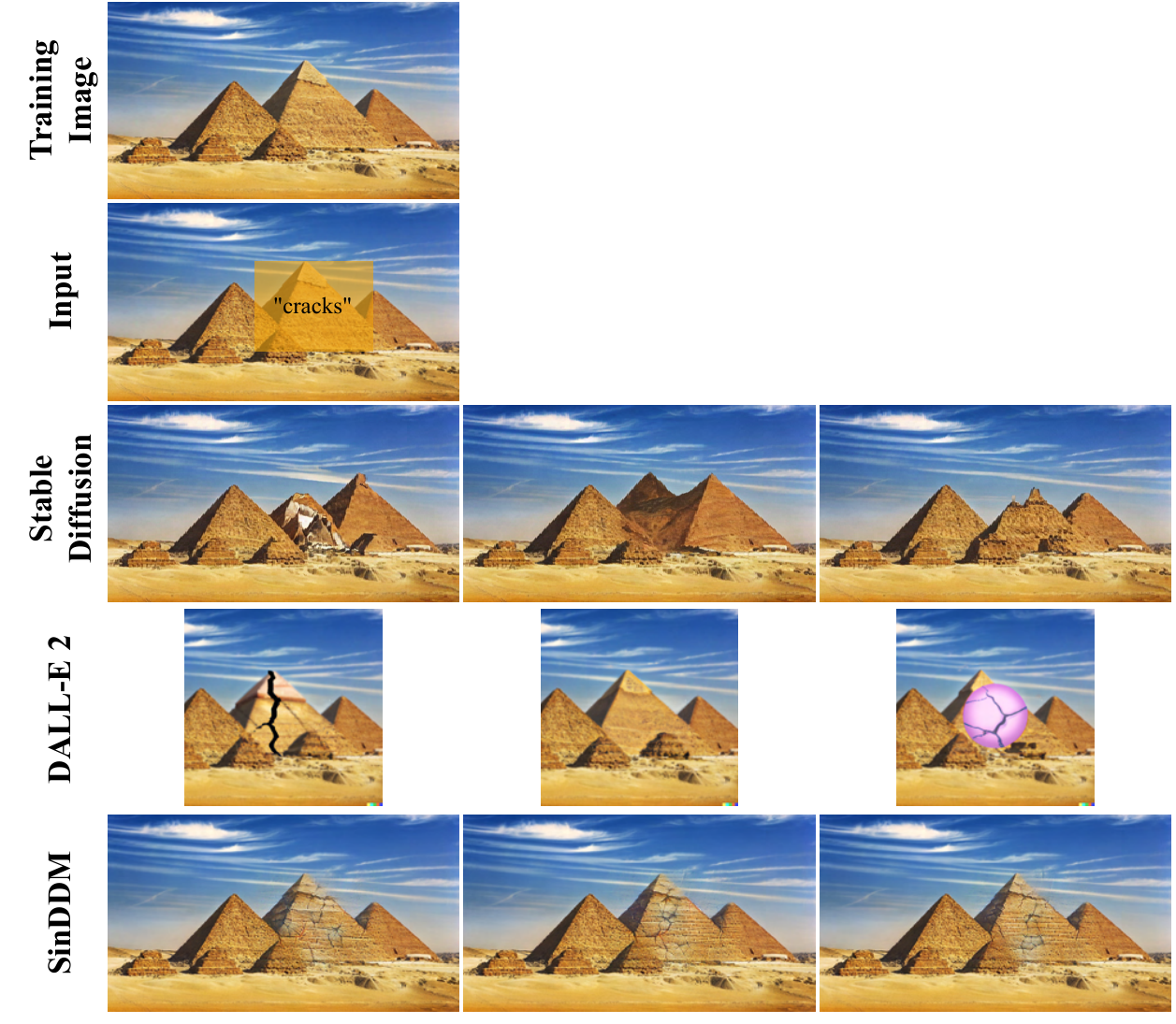}
  \caption{\textbf{Comparison of text-guided content generation in ROI.} We show comparisons to the Stable Diffusion inpainting and DALL-E~2 editing methods on the task of text-guided generation in a ROI.}
  \label{fig:clip_roi_comp}
\end{figure}

In \cref{fig:clip_roi_comp} we compare our text-guided content generation in ROI to the Stable Diffusion inpainting method (as implemented in HuggingFace diffusers\footnote{\url{https://huggingface.co/docs/diffusers/using-diffusers/inpaint}}) and to the DALL-E 2 editing option that is accessible via their API\footnote{\url{https://openai.com/api/}}. 
The Stable Diffusion inpainting method creates images that are not loyal to the text input and and do not blend well with the original image. Using the DALL-E~2 API, the image needed to be cropped and reshaped to $1024\times1024$ prior to editing. As can be seen, the DALL-E 2 results are unrealistic and in some cases the results are not related to the input text (like in the middle example, where the DALL-E~2 results do not depict cracks). 

\clearpage
\subsection{Generation with Text-Guided Style}\label{appendix:comp_style}
\begin{figure}[H]
  \noindent
  \includegraphics[width=\textwidth]{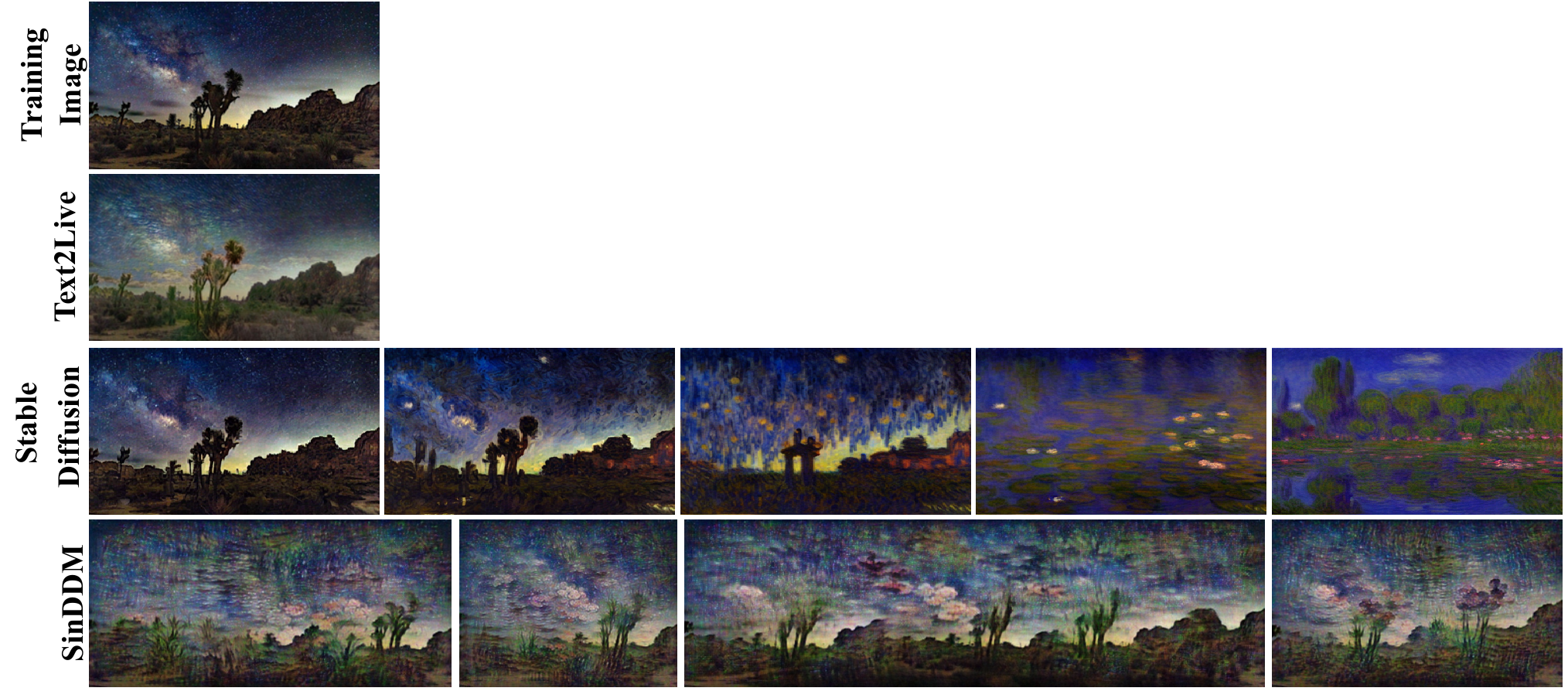}
  \caption{\textbf{Comparisons for image generation with text-guided style.} Here we used the text prompt ``Monet style''. For Text2Live we provided ``Monet style'' as the text describing the edit layer, ``a desert with vegetation in night under starry sky in the style of Monet'' as the text describing the full edited image and ``a desert with vegetation in night under starry sky'' as the text describing the input image and as the text describing the region of interest in the input image. For Stable Diffusion we show strengths from 0.2 (left) to 1 (right) in jumps of 0.2.}
  \label{fig:style_comp}
\end{figure}

\begin{figure}[H]
  \noindent
  \includegraphics[width=\textwidth]{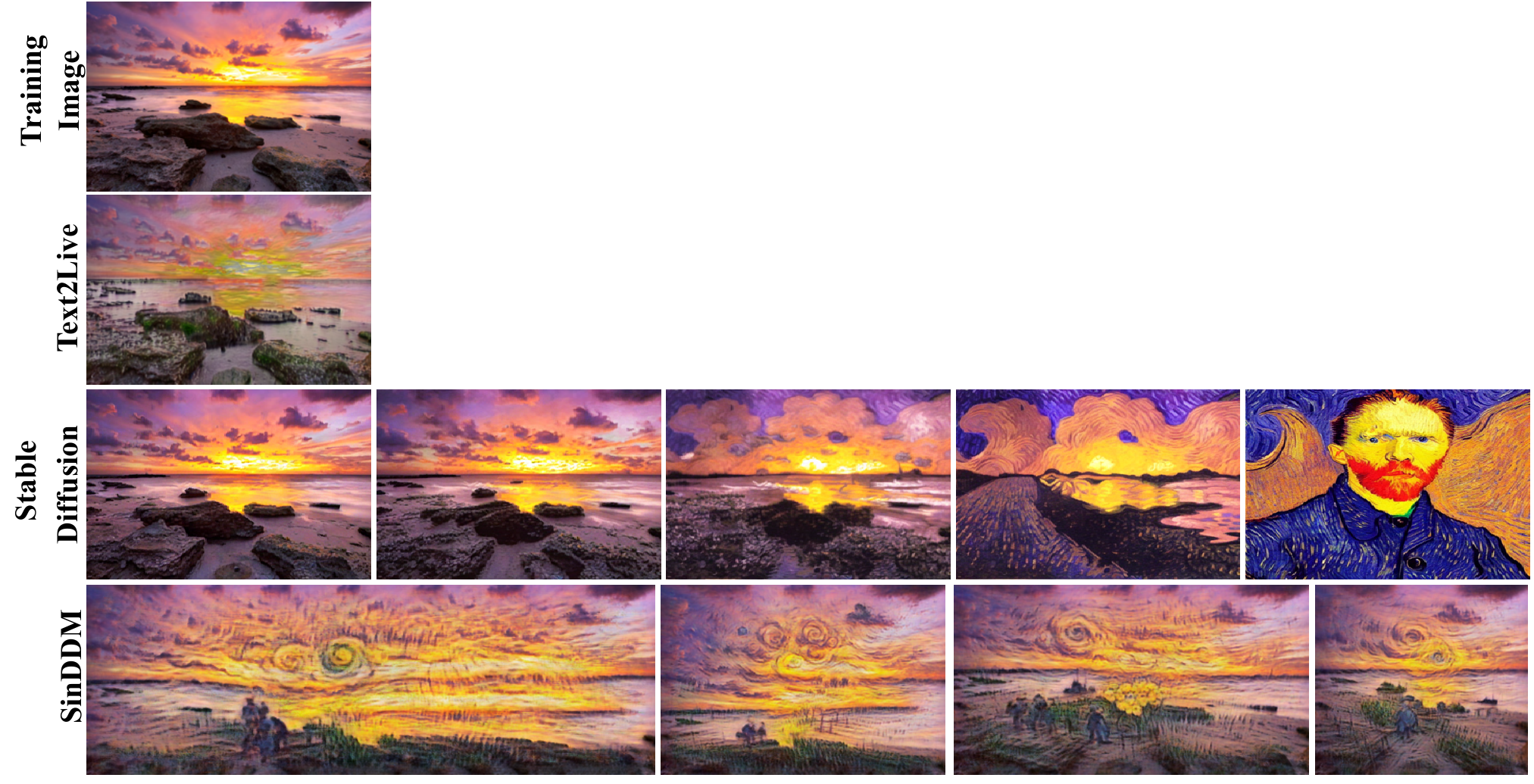}
  \caption{\textbf{Comparisons for image generation with text-guided style.} Here we used the text prompt ``Van Gogh style''. For Text2Live we provided ``Van Gogh style'' as the text describing the edit layer, ``an isolated sea shore with rocks and white sand, during pink and orange sunset in the style of Van Gogh'' as the text describing the full edited image and ``an isolated sea shore with rocks and white sand, during pink and orange sunset'' as the text describing the input image and as the text describing the region of interest in the input image. For Stable Diffusion we show strengths from 0.2 (left) to 1 (right) in jumps of 0.2.}
  \label{fig:style_comp1}
\end{figure}

\begin{figure}[H]
  \noindent
  \includegraphics[width=\textwidth]{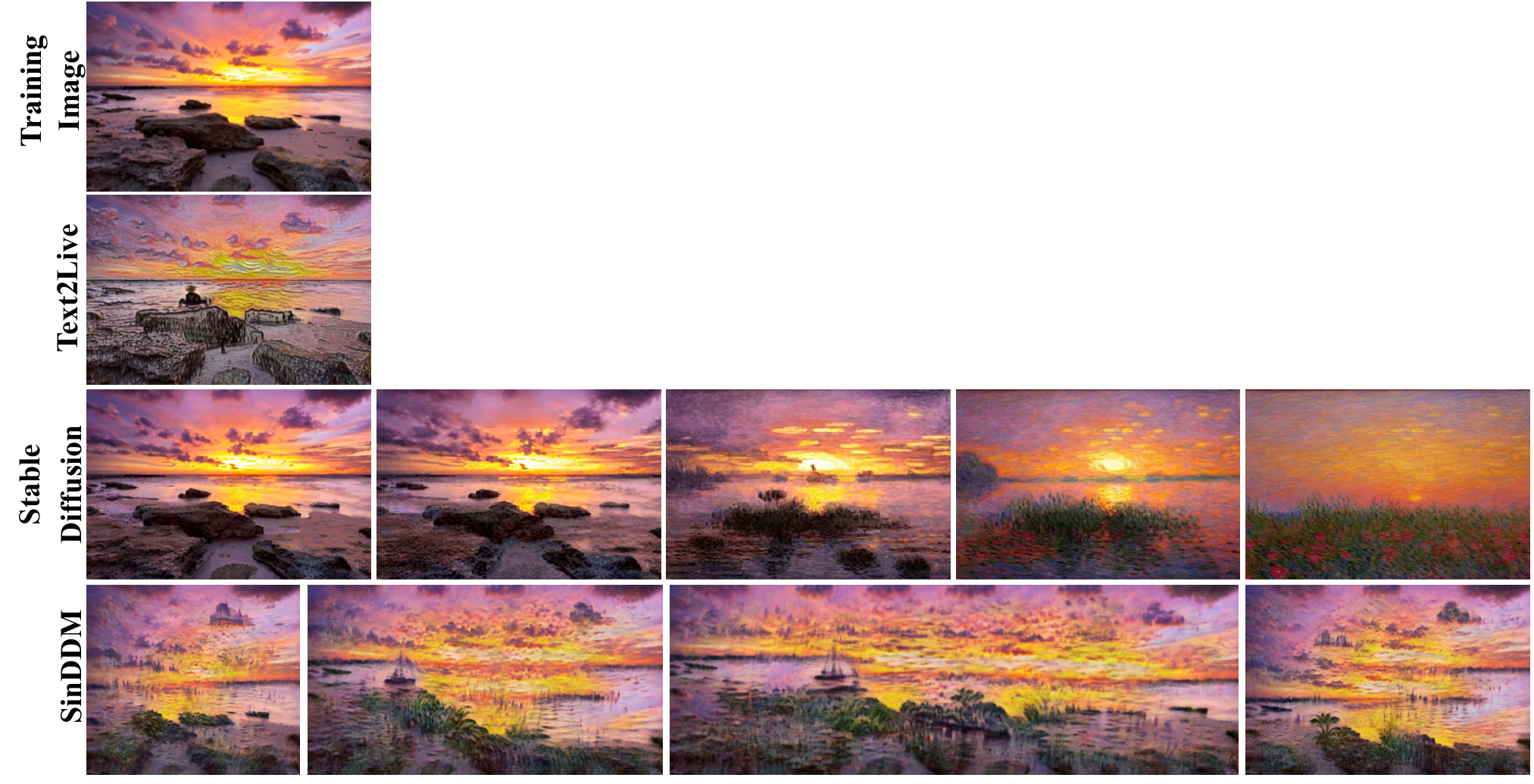}
  \caption{\textbf{Comparison of image generation guided by style.} Here we used the text prompt ``Monet style''. For Text2Live we provided the same prompts as described in \cref{fig:style_comp1} but used ``Monet'' instead of ``Van Gogh''. For Stable Diffusion we show strengths from 0.2 (left) to 1 (right) in jumps of 0.2.}
  \label{fig:style_comp2}
\end{figure}

Here we compare our method of generation with text-guided style to Text2Live and Stable Diffusion. The comparisons are shown in Figures~\ref{fig:style_comp}, \ref{fig:style_comp1} and \ref{fig:style_comp2}.
\paragraph{Text2Live} As in App.~\ref{appendix:comp_content}, for Text2Live, we used 1000 bootstrap epochs. Furthermore, this method requires four text prompts (three prompts on top of the one for describing the final desired result, as in our method). In each example, we list all four prompts.
\paragraph{Stable Diffusion} For Stable Diffusion, we used our modified version of the image-to-image option that reshapes the image to $512\times512$ before injecting it to the Stable Diffusion pipeline. 
In each example we show strengths from 0.2 (left) to 1 (right) in jumps of $0.2$.
As before, when the strength parameter is small, the edited image is loyal to the original image but the effect of the text is barely noticeable. 
When the strength parameter is large, the effect of the text is substantial, but the result is no longer similar to the original image.

\clearpage

\clearpage
\subsection{Style Transfer}\label{appendix:style_transfer}
\begin{figure}[H]
  \noindent
  \includegraphics[width=\textwidth]{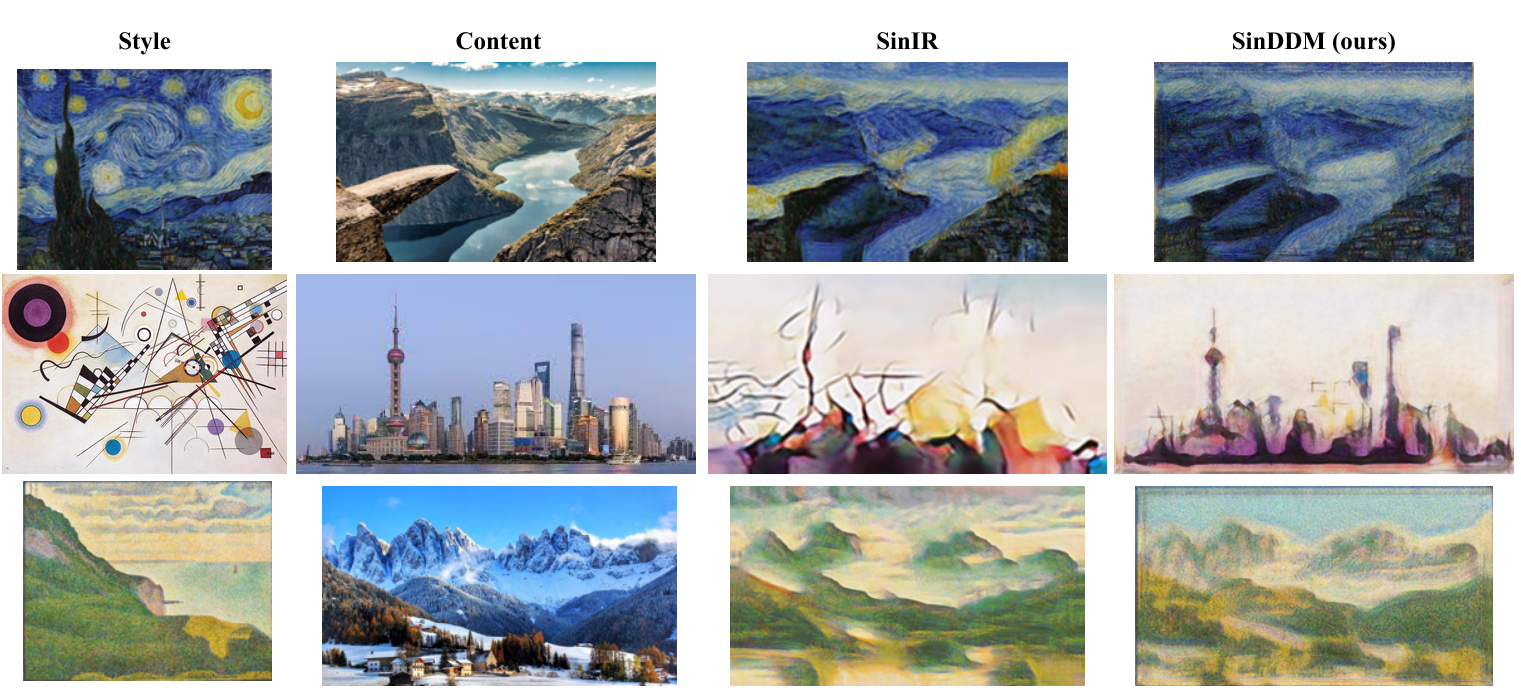}
  \caption{\textbf{Style Transfer.} Comparison between SinIR and SinDDM (our).}
  \label{fig:style}
\end{figure}

In \cref{fig:style}, we compare SinDDM to SinIR on the task of style transfer. As can be seen, SinDDM can generate results that are simultaneously more loyal to the content image and to the style image.

\clearpage

\subsection{Harmonization}\label{appendix:harmo}
\begin{figure}[H]
  \noindent
  \includegraphics[width=\textwidth]{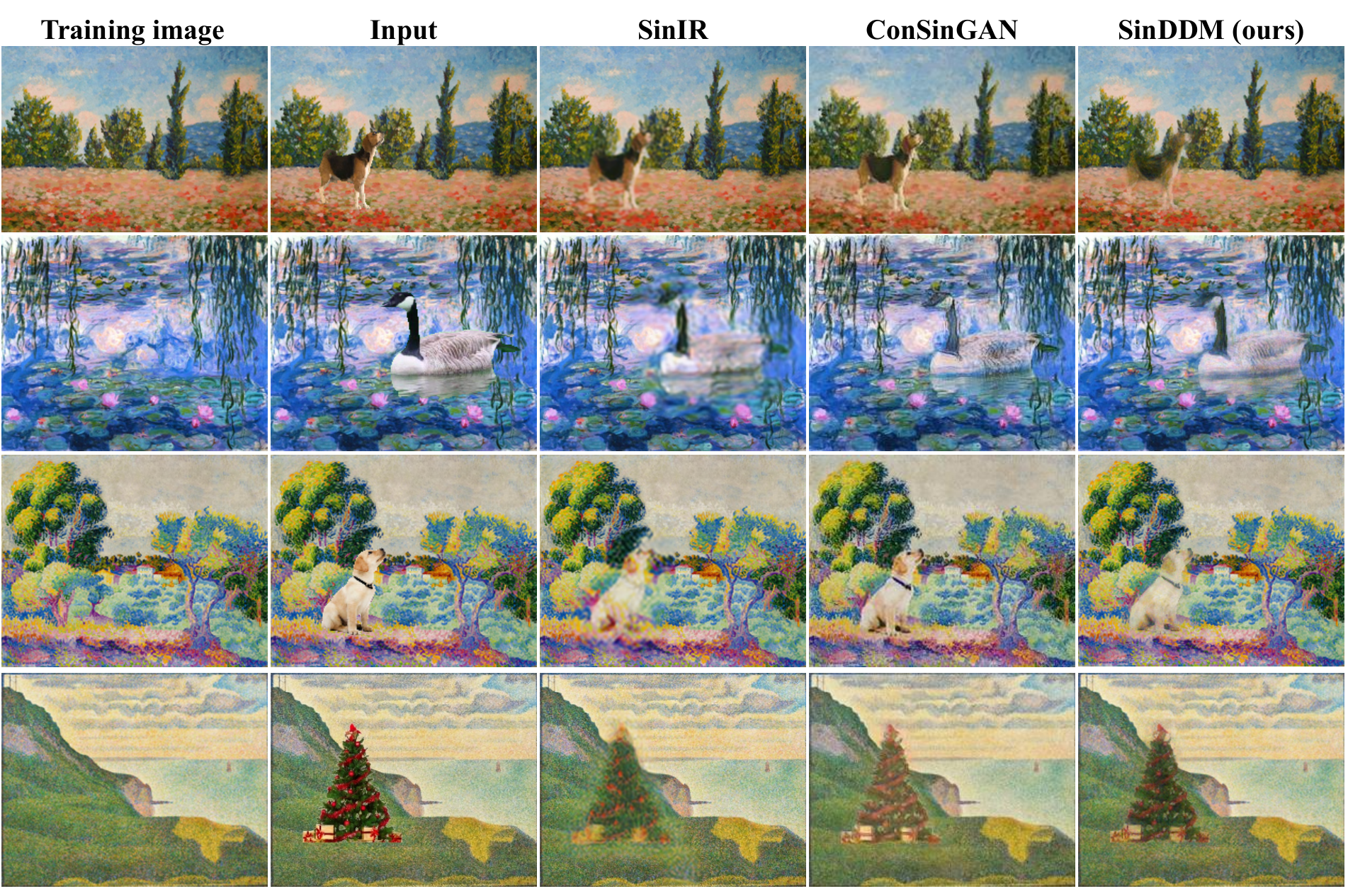}
  \caption{\textbf{Harmonization.} Comparison between SinIR, ConSinGAN and SinDDM (our). For ConSinGAN, we used the fine-tuning option provided in the code, which fine-tunes the trained model on the naively pasted image.}
  \label{fig:harmo}
\end{figure}

In \cref{fig:harmo}, we compare SinDDM to SinIR and ConSinGAN on the task of harmonization. Here SinDDM leads to stronger blending effects, but at the cost of some blur in the pasted object.

\clearpage
\section{Limitations and Directions for Future Work}\label{appendix:FutureWork}

\subsection{Misrepresentation of the Distribution of Certain Objects}\label{appendix:misrepresentation}
\begin{figure}[H]
  \centering
  \includegraphics[width=\textwidth]{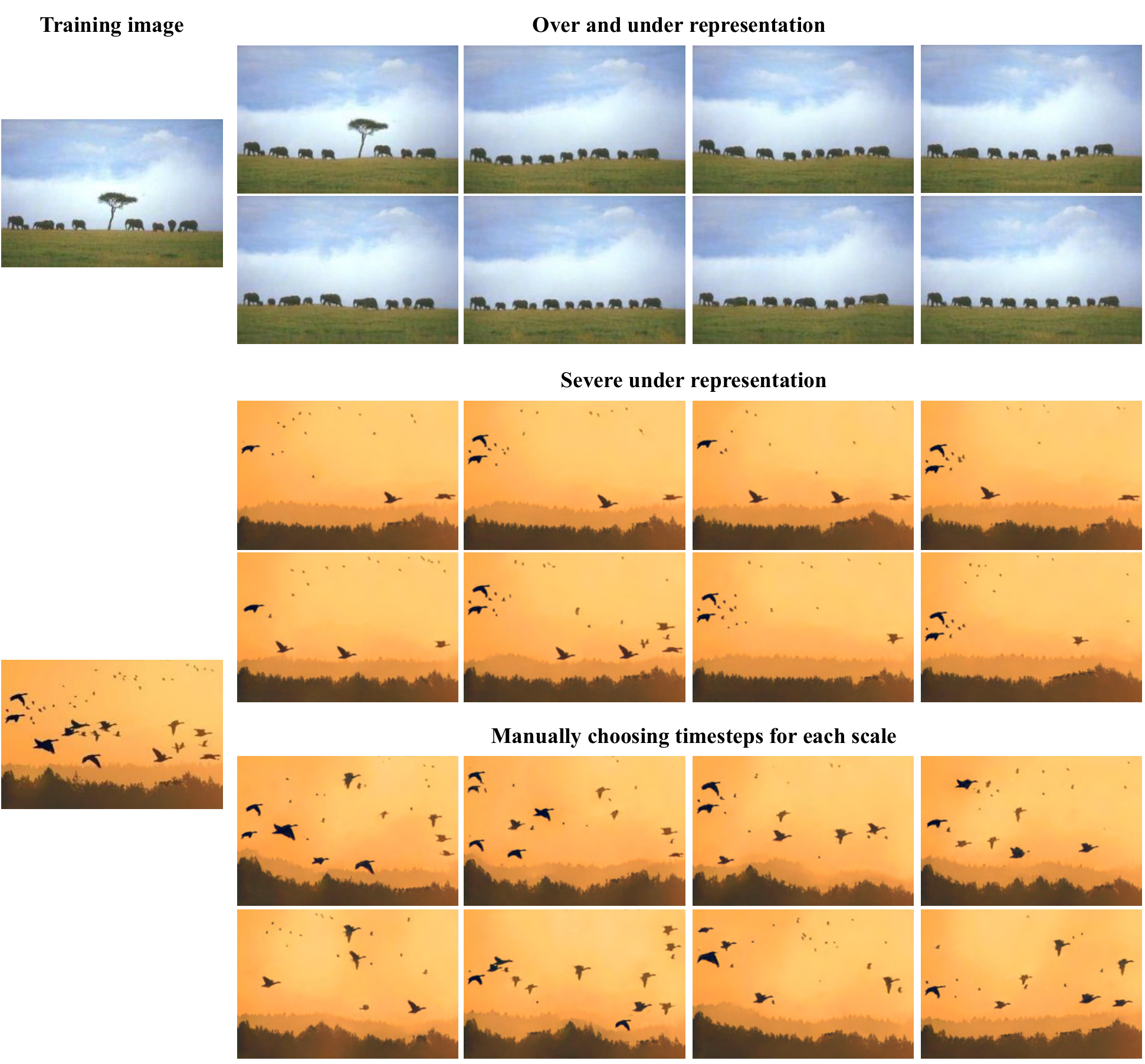}
  \caption{\textbf{Limitations.} Our model can sometimes under- or over-represent certain objects in the training image, as seen in the Elephants and Birds images. These problems can be fixed by manually choosing the initial timestep for each scale. However we do not currently have an automatic way to choose these values, which avoids these problems for all training images.}
  \label{fig:ovunrepr}
\end{figure}

Our sampling algorithm often misrepresents the true distribution of the objects in the original image. For example, in the Elephants image in \cref{fig:ovunrepr}, our samples usually contain more elephants than in the original image, and less trees (more often than not, the samples do not contain a single tree). In certain images with many small objects, our sampling algorithm can fail to represent them faithfully. An extreme example is the Birds image in \cref{fig:ovunrepr}, where the birds are severely under-represented in our samples. This can be solved by (1) manually tuning the initial $T$ for each scale, and (2) modifying the sampling scheme such that it does not involve the blending with the image generated at the previous scale. Such a fix is illustrated at the bottom of \cref{fig:ovunrepr}. However, for (1) we currently do not have an automatic algorithm which works well across all images, and using~(2) introduces undesired blur to the samples. We therefore leave these directions for future research.

\subsection{Inner Distribution Preservation Under Text Guided Content Generation}\label{appendix:inner_dist}
An additional limitation of our method relates to text guidance. Specifically, when using text guidance to generate an image with new content, we are limited to images that adhere to the internal distribution of the training image. This limits the new content to contain only textures which exist in the training image. For example, ``fire in the forest'' seen in \cref{fig:clip_comp_sup3} looks realistic because the orange sky texture was used to create the fire.

\subsection{Boundary conditions and the effect of padding}
\begin{figure}[H]
  \centering
  \includegraphics[width=\textwidth]{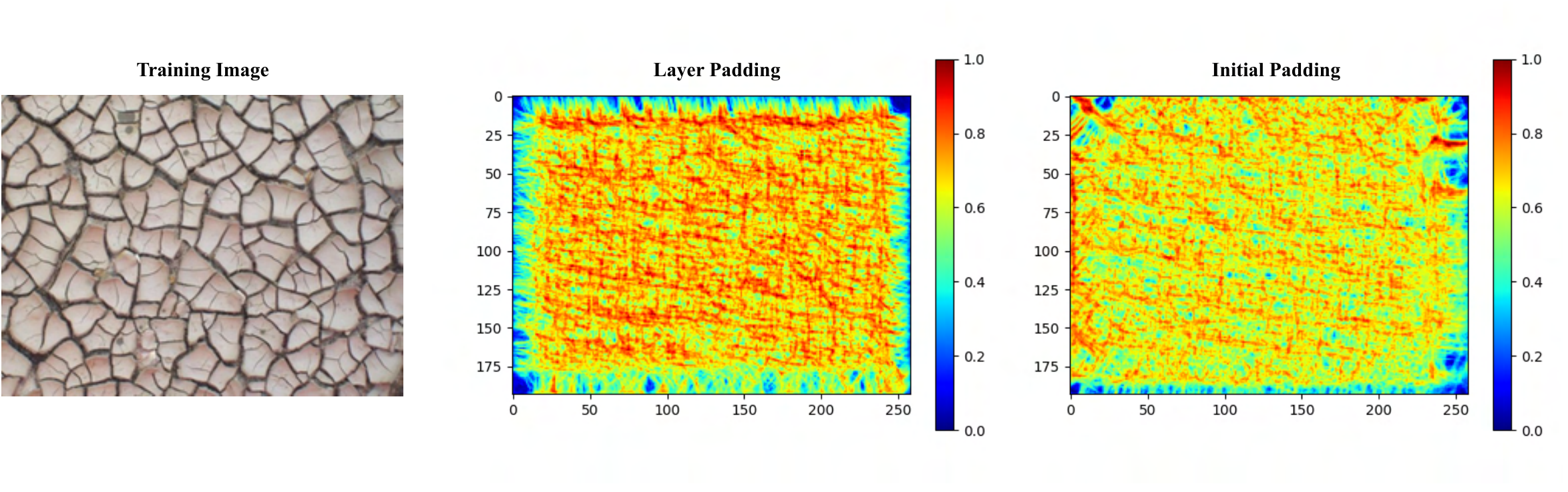}
  \caption{\textbf{Effect of padding.} The padding configuration affects the diversity between samples at the corners. Zero padding in each convolutional layer (layer padding) leads to small variability at the borders but large variability away from the borders. Padding only the input to the net (initial padding), leads to increased variability at the borders but smaller variability at the interior part of the image.}
  \label{fig:padding_effect}
\end{figure}

Similarly to SinGAN (see Supplementary Sec.~2 in \cite{shaham2019singan}), the type of padding used in the model influences the diversity among generated samples at the corners. Figure~\ref{fig:padding_effect} illustrates this effect by depicting the standard deviation among generated samples for each location in the image. As can be seen, zero padding in each convolutional layer (layer padding) leads to small variability at the borders but large variability away from the borders. Padding only the input to the network (initial padding), leads to increased variability at the borders but smaller variability at the interior part of the image

\clearpage

